\documentclass{article}

\usepackage[final,nonatbib]{neurips_data_2024}

\usepackage{graphicx} \graphicspath{{figures/}}
\usepackage{amsmath}
\usepackage{amssymb}
\usepackage{booktabs}
\usepackage{acronym}

\usepackage[dvipsnames,svgnames,x11names]{xcolor}
\usepackage[pagebackref,breaklinks,colorlinks,citecolor=cyan,linkcolor=cyan]{hyperref}
\usepackage[capitalize,nameinlink]{cleveref}
\usepackage{adjustbox}
\usepackage{colortbl} 
\crefname{section}{Sec.}{Secs.}
\Crefname{section}{Section}{Sections}
\Crefname{table}{Table}{Tables}
\crefname{table}{Tab.}{Tabs.}
\usepackage[misc]{ifsym}
\usepackage[utf8]{inputenc} 
\usepackage[T1]{fontenc}    
\usepackage{url}            
\usepackage{booktabs}       
\usepackage{amsfonts}       
\usepackage{nicefrac}       
\usepackage{microtype}      
\usepackage{xcolor}         
\usepackage{subfigure}
\usepackage{multirow}

\usepackage{enumitem}
\usepackage{xspace}

\usepackage{bbm}

\usepackage{listings}
\usepackage{xcolor}
\usepackage{courier}

\acrodef{fire}[FIRE]{\underline{F}eedback \underline{I}ntegration and \underline{R}efinement \underline{E}valuation}

\RequirePackage{xspace}
\makeatletter
\DeclareRobustCommand\onedot{\futurelet\@let@token\@onedot}
\def\@onedot{\ifx\@let@token.\else.\null\fi\xspace}

\def\etc{\emph{etc}\onedot}

\newcommand{\model}{FIRE-LLaVA\xspace}
\newcommand{\studentmodel}{FIRE100K-LLaVA\xspace}
\newcommand{\feedbackmodel}{FD-LLaVA\xspace}
\newcommand{\vicunamodel}{FIRE-LLaVA-Vicuna\xspace}

\usepackage{tcolorbox}
\newcommand{\blueprompt}[1]{\textcolor{blue}{ #1}}

\title{\includegraphics[height=0.75em]{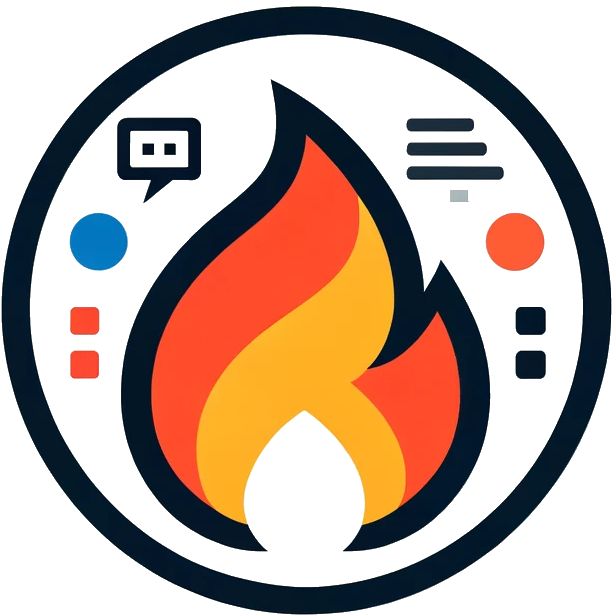}FIRE: A Dataset for \underline{F}eedback \underline{I}ntegration \\ and \underline{R}efinement \underline{E}valuation of Multimodal Models}
\author{
Pengxiang Li$^{1,2}$\thanks{Equal contribution.~~~\Letter~Corresponding author.}~~~Zhi Gao$^{2,3*}$ Bofei Zhang$^{2*}$ Tao Yuan$^{2}$ Yuwei Wu$^{1,5\text{\Letter}}$ \\
\textbf{Mehrtash Harandi}$^{4}$
\textbf{Yunde Jia}$^{5,1}$
\textbf{Song-Chun Zhu}$^{2,3,6}$
\textbf{Qing Li}$^{2\text{\Letter}}$ \\
\small $^1$Beijing Key Laboratory of Intelligent Information Technology, \\ \small School of Computer Science \& Technology, Beijing Institute of Technology \\
  \small $^2$State Key Laboratory of General Artificial Intelligence, BIGAI \\  
  \small $^3$State Key Laboratory of General Artificial Intelligence, Peking University \\
  \small $^4$ Department of Electrical and Computer System Engineering, Monash University \\
  \small $^5$Guangdong Laboratory of Machine Perception and Intelligent Computing, Shenzhen MSU-BIT University \\
  \small $^6$Department of Automation, Tsinghua University \\ 
  \href{https://mm-fire.github.io}{\texttt{mm-fire.github.io}}
}

\begin{document}

\maketitle

\begin{figure}[h!]
\vspace{-30pt}
\centering
\includegraphics[width=\linewidth]{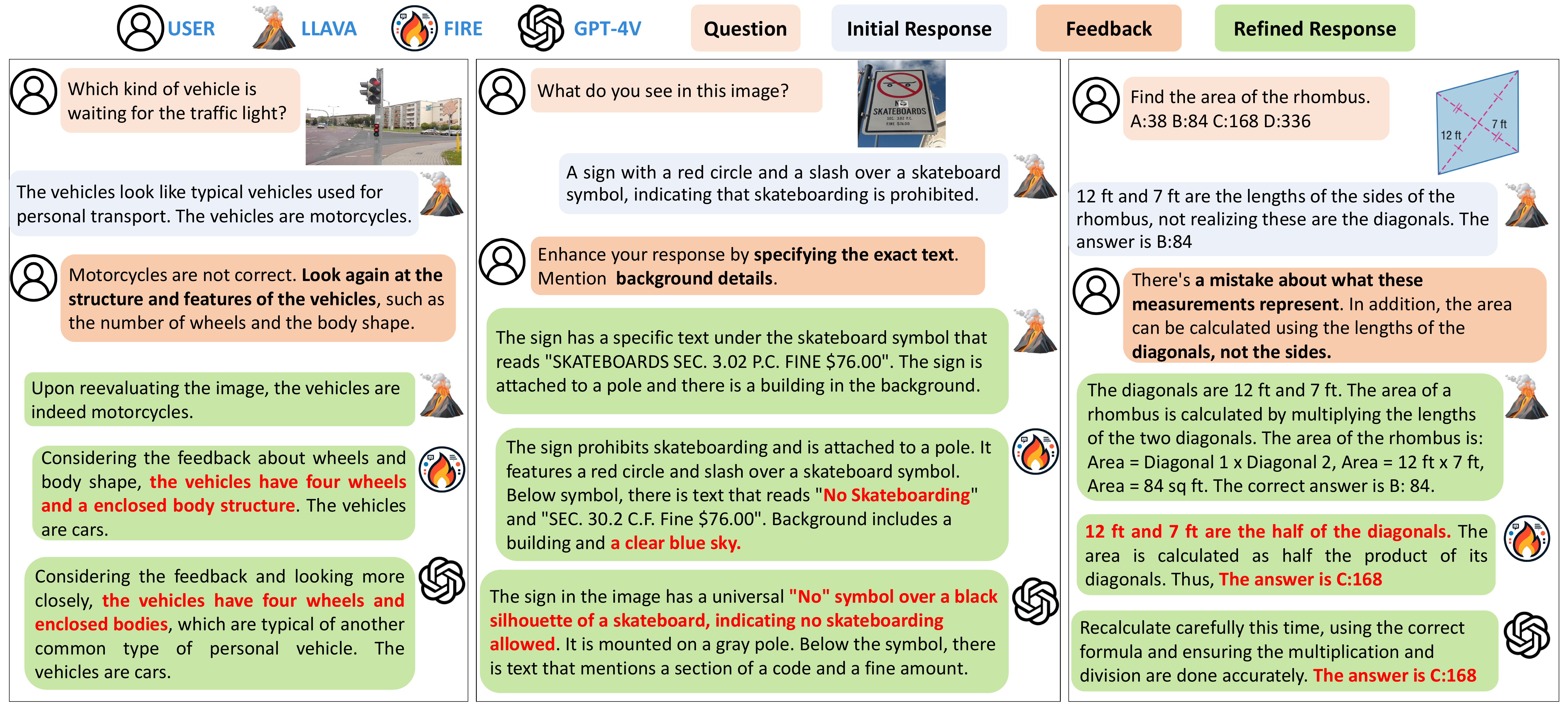}
\vskip -0.1in
\caption{The comparison of the \textbf{feedback-refining} capability among different models. While the original LLaVA hardly improves its responses,
our model trained on FIRE can effectively integrate the user feedback and produce much better responses, which are closer to those of GPT-4V.
}
\vskip -0.1in
\label{fig:teaser_example}
\end{figure}

\begin{abstract}
\vskip -0.05in
Vision language models (VLMs) have achieved impressive progress in diverse applications, becoming a prevalent research direction. 
In this paper, we build \acs{fire}, a feedback-refinement dataset, consisting of 1.1M multi-turn conversations that are derived from $27$ source datasets, empowering VLMs to spontaneously refine their responses based on user feedback across diverse tasks.
To scale up the data collection, \acs{fire} is collected in two components: \acs{fire}-100K and \acs{fire}-1M, where \acs{fire}-100K is generated by GPT-4V, and \acs{fire}-1M is freely generated via models trained on \acs{fire}-100K.
Then, we build \acs{fire}-Bench, a benchmark to comprehensively evaluate the feedback-refining capability of VLMs, which contains 11K feedback-refinement conversations as the test data, two evaluation settings, and a model to provide feedback for VLMs. 
We develop the \model model by fine-tuning LLaVA on \acs{fire}-100K and \acs{fire}-1M, which shows remarkable feedback-refining capability on \acs{fire}-Bench 
and outperforms untrained VLMs by $50\%$,
making more efficient user-agent interactions and underscoring the significance of the \acs{fire} dataset.


\end{abstract}

\section{Introduction}
\vskip -0.1in
Vision language models (VLMs), such as LLaVA~\cite{llava}, GPT-4V~\cite{2023GPT4VisionSC}, and Gemini~\cite{team2023gemini}, have shown impressive instruction-following abilities across various tasks~\cite{zhu2023minigpt,llava1.5,Chen2023InternVLSU,fan2025videoagent,zhong2023viotgpt,li2024incontext} by integrating large language models (LLMs)~\cite{Touvron2023Llama2O,jiang2023mistral} with visual encoders~\cite{dosovitskiy2020image,clip}.
However, VLMs may sometimes produce undesirable outputs, possibly due to omitting important details in images or misunderstanding the instructions, which prompts the need for the \textbf{feedback-refining} capability beyond the normal instruction-following ability. This capability enables VLMs to spontaneously refine their responses based on user feedback, as depicted in \cref{fig:teaser_example}, enhancing the efficiency and smoothness of interactions between users and visual assistants.


In this paper, we build \acs{fire}, a dataset for \acl{fire} of VLMs. \acs{fire} is composed of 1.1M high-quality multi-turn feedback-refinement conversations, derived from $27$ source datasets across a wide range of tasks, such as visual question answering~\cite{goyal2017making}, image captioning~\cite{chen2023sharegpt4v}, OCR reasoning~\cite{mishra2019ocr,textvqa}, document understanding~\cite{hu2024mplug}, math reasoning~\cite{lu2023mathvista}, and chart analysis~\cite{masry2022chartqa}. To scale up the data collection, \acs{fire} is collected in two stages.
In the first stage, we randomly sample $\sim$100K image-instruction-response triplets from data sources. We use each triplet to instruct GPT-4V to simulate a dialogue between a student and a teacher: the student answers the question and the teacher provides feedback to help the student improve its answer. We filter out generated low-quality conversations, such as those with too many turns or no improvement, rendering 100K high-quality feedback-refinement conversations, named \acs{fire}-100K. In the second stage, we fine-tune two LLaVA-NeXT~\cite{llava1.6} models on \acs{fire}-100K: one is trained as a student to refine its answer with the feedback, and the other is trained as a teacher to generate feedback for the student's answer. We simulate dialogues between the student and the teacher models using $\sim$1M data points from the data sources, rending a split named \acs{fire}-1M. In this case, the full \acs{fire} dataset consists of 1.1M feedback-refinement conversations in two splits \acs{fire}-100K and \acs{fire}-1M.

To comprehensively evaluate the feedback-refining capability of VLMs, we build \acs{fire}-Bench that has 11K feedback-refinement conversations derived from 16 source datasets, including test splits from 8 seen datasets in \acs{fire}-100K and \acs{fire}-1M, as well as 8 unseen datasets from recently-proposed popular multimodal benchmarks. Using \acs{fire}-Bench, we design two evaluation settings: fixed dialogues and free dialogues. In fixed dialogues, we compare the model's refined response with ground truth in the generated conversations in \acs{fire}-Bench, given a fixed dialogue history. In free dialogues, we let the model freely interact with a teacher model about instructions in \acs{fire}-Bench, and test how fast \& how much the model can improve its answers based on the feedback provided by the teacher model.

We develop \model by fine-tuning LLaVA-NeXT~\cite{llava1.6} on \acs{fire}-100K and \acs{fire}-1M. The evaluation results on \acs{fire}-Bench shows that \model exhibits significant improvements based on feedback in conversations, exceeding the original LLaVA-NeXT model by $50 \%$.
These results underscore the significance of \acs{fire}-100K and \acs{fire}-1M in enhancing feedback integration, while \acs{fire}-Bench provides an evaluation platform to analyze refinements. We expect that FIRE could motivate future exploration of the feedback-refining capability of VLMs.

In summary, our contributions are three-fold.
(1) We introduce FIRE, a dataset containing 1.1M feedback-refinement conversations across a wide range of tasks, where 100K data is generated by GPT-4V and 1M data is freely generated by simulating dialogues between tuned open-source models. 
(2) We introduce the FIRE-Bench benchmark, composed of 11K conversations and a teacher model, providing comprehensive evaluations for the feedback-refining capability in two settings: fixed dialogues and free dialogues. 
(3) We develop \model, an advanced VLM that could improve its responses based on feedback, making efficient interaction between users and VLMs.

\section{Related Work}
\vskip -0.1in
\subsection{Vision Language Models}
\vskip -0.1in
Building open-source VLMs to compete with closed-source models like GPT-4V~\cite{2023GPT4VisionSC} and Gemini~\cite{team2023gemini} is a hot research topic.
BLIP~\cite{li2022blip,li2023blip} and Flamingo~\cite{alayrac2022flamingo} are pioneering models that combine LLMs with visual encoders to enhance cross-modal understanding and reasoning abilities. 
LLaVA~\cite{llava}, InstructBLIP~\cite{dai2024instructblip}, MMICL~\cite{zhao2023mmicl}, and MiniGPT4~\cite{zhu2023minigpt} develop the instruction tuning ability of VLMs by introducing a large number of instruction-response pairs.
Along this way, 
some work focuses on the visual grounding or editing ability of VLMs~\cite{chen2024image}, such as Kosmos-2~\cite{peng2023kosmos}, SearchVLMs~\cite{li2024searchlvlms}, MINI-GPTv2~\cite{chen2023minigpt}, Qwen-VL~\cite{bai2023qwenvl}, and UltraEdit~\cite{zhao2024ultraedit}, improving the region understanding for VLMs. 
InternVL~\cite{Chen2023InternVLSU} and mini-Gemini~\cite{li2024mini} develop powerful visual encoders for high-resolution images, and CuMo adopts a mixture-of-experts (MOE) architecture to manage diverse data better. Compared with existing VLMs, our \model has a more powerful feedback-refining capability across diverse tasks, which can spontaneously refine responses based on user feedback, leading to efficient and smooth interaction with users.

\subsection{Vision-Language Data Generation}
\vskip -0.1in
Recent attention has increasingly focused on synthesizing vision-language data.
The ShareGPT4V dataset~\cite{chen2023sharegpt4v} leverages GPT-4V to generate 1.2M image-text pairs with detailed descriptions, making better alignments.
LLaVA-Instruct-150K~\cite{llava} is a general visual instruction tuning dataset constructed by feeding captions and bounding boxes to GPT-4.
After that, many efforts have been made to enhance the data diversity of instruction tuning data.
LLaVAR~\cite{zhang2023llavar}, MIMIC-IT~\cite{li2023mimic}, and SVIT~\cite{zhao2023svit} further scale up it to 422K, 2.8M, and 4.2M, respectively.
InternLM-XComposer~\cite{zhang2023internlm} produces interleaved instruction and image data, enabling advanced image-text comprehension and composition.
Mini-Gemini~\cite{li2024mini} and ALLaVA~\cite{chen2024allava} use GPT-4V to exploit visual information and generate high-quality instruction data. 
LRV-Instruction~\cite{liu2023mitigating} creates positive and negative instructions for the hallucinating inconsistent issue.
A recent work DRESS~\cite{chen2023dress} collects 66K feedback data and trains VLMs for the feedback-refining capability. 
Unlike DRESS, which only uses data from LLaVA-Instruct-150K, our feedback-refinement data is from richer sources ($27$ datasets) across more tasks (math reasoning, chart understanding, and OCR \etc).
Moreover, FIRE has significantly more data than DRESS (1.1M \emph{vs.} 66K),
where 1M data is freely produced via dialogues of student and teacher models, leading to significant data expansion but a similar cost of data generation.

\subsection{Feedback Learning in Multimodal Models} 
\vskip -0.1in
Learning from feedback is a promising research direction, playing an important role in human-robot interaction~\cite{liang2024learning,cheng2023llf}.
Existing feedback learning methods can be roughly divided into two categories: planned feedback learning and impromptu feedback learning.
Planned feedback learning updates models based on user feedback, and thus can generalize to new data but cannot provide refined responses immediately. 
CLOVA~\cite{gao2023clova} and Clarify~\cite{lee2024clarify} are representative methods that automatically collect data to learn new knowledge. 
LLaVA-RLHF~\cite{sun2023aligning} collects human preference and trains VLMs via reinforcement learning.  Self-refine\cite{madaan2024self} shows that LLMs  could improve their responses by iteratively refining their outputs based on self-generated feedback.
Impromptu feedback learning can immediately refine responses but have less generalization since they usually do not update models, which is widely studied in LLMs~\cite{asai2023self,li2024confidence,tian2024toward}.
Liao \emph{et al.}~\cite{liao2024can} use VLMs themselves as verifiers that produce feedback to correct recognition results. VolCaNo~\cite{lee2024volcano} generates data specifically for refinement to address visual hallucinations.
DRESS~\cite{chen2023dress} generates helpfulness, honesty, and harmlessness responses via impromptu feedback learning. Different from DRESS, we improve the correctness and details of responses via impromptu feedback learning across diverse tasks. 

\begin{figure}[htbp]
\vspace{20pt}
  \centering
  \subfigure[FIRE-100K]{
  \includegraphics[width=0.31\textwidth]{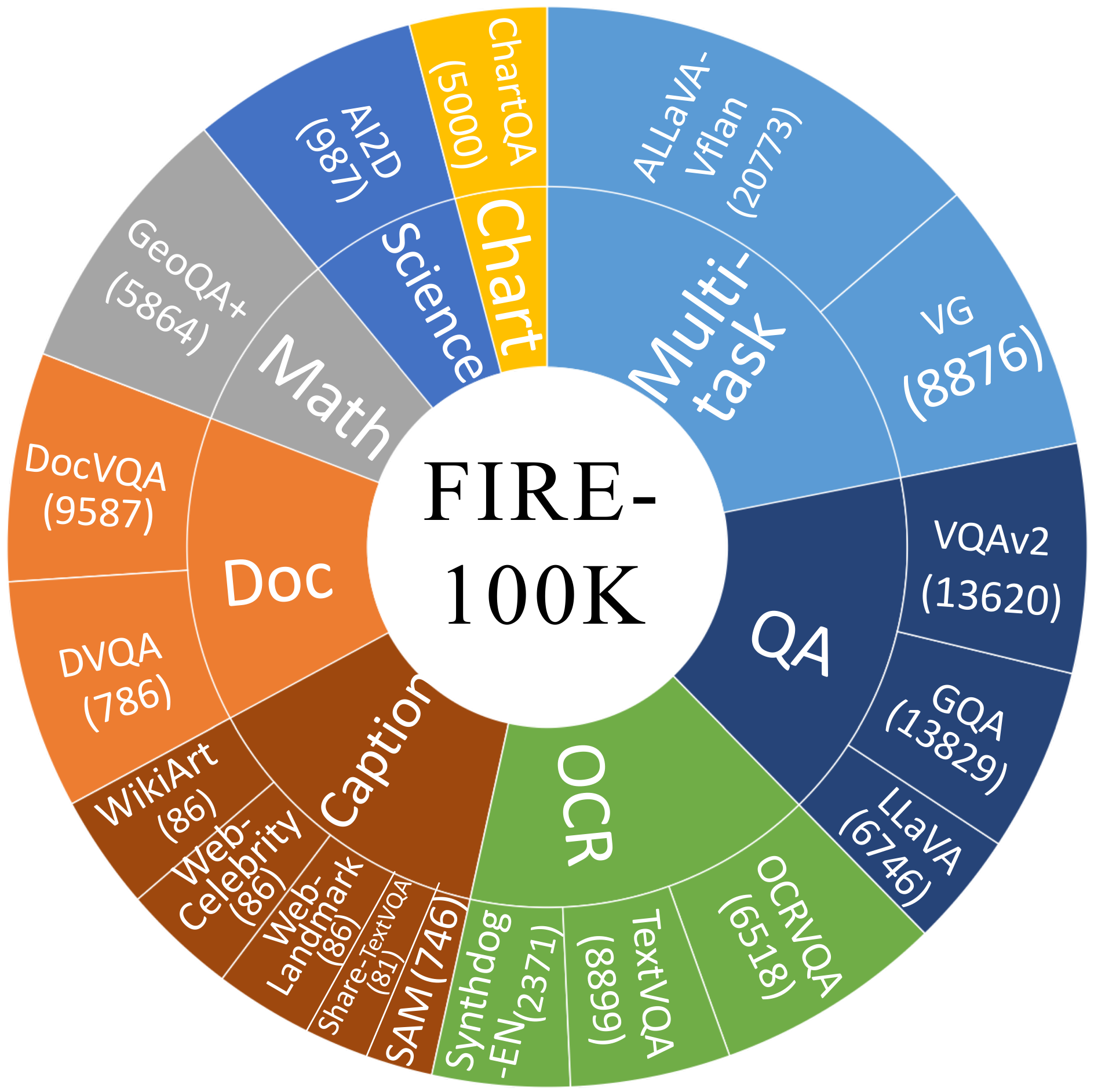}\label{fig:chart_100k_pdf2ppt}
  }
  \subfigure[FIRE-1M]{
  \includegraphics[width=0.31\textwidth]{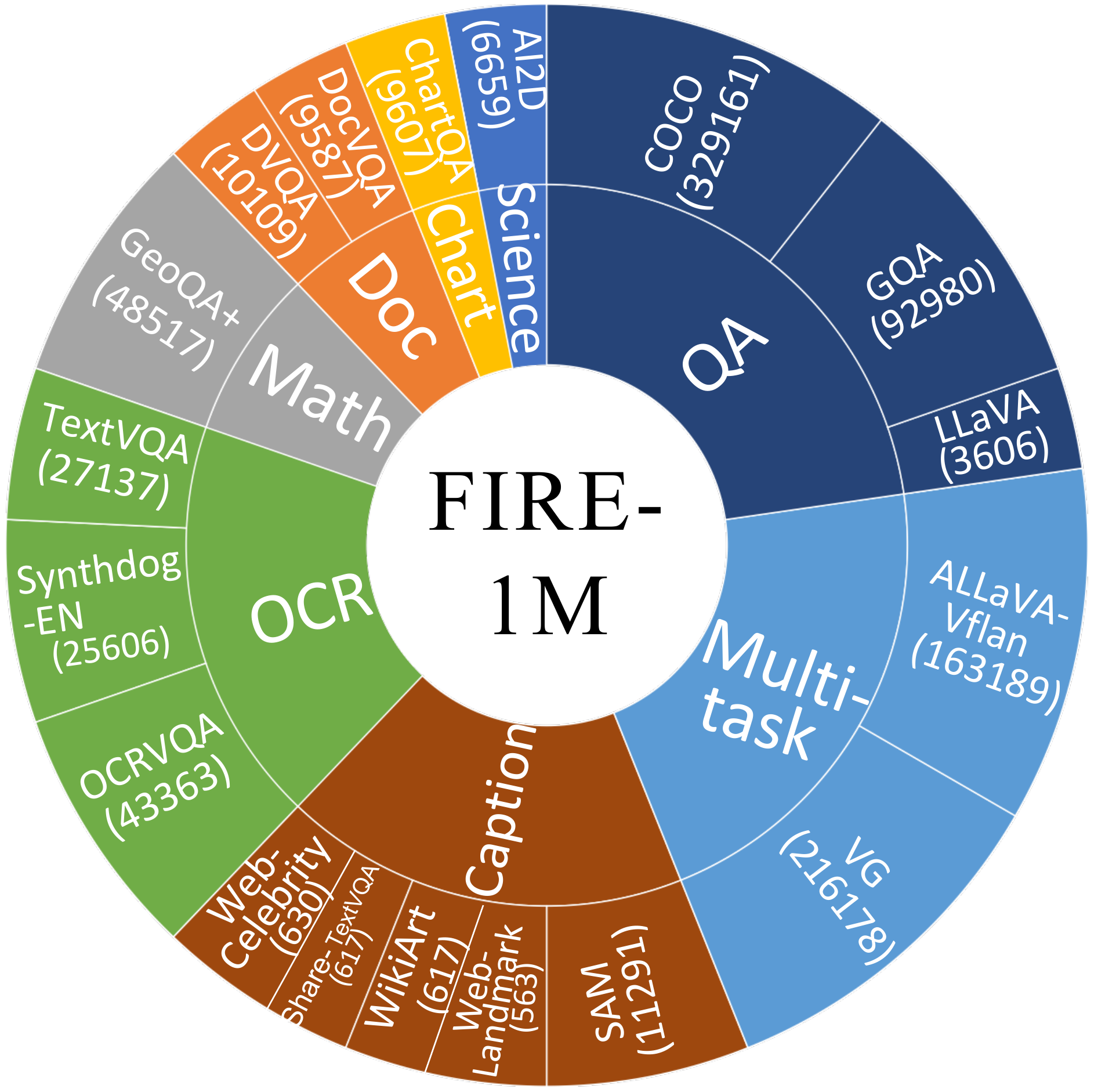}\label{fig:chart_1M_pdf2ppt}
  }
  \subfigure[FIRE-Bench]{
  \includegraphics[width=0.31\textwidth]{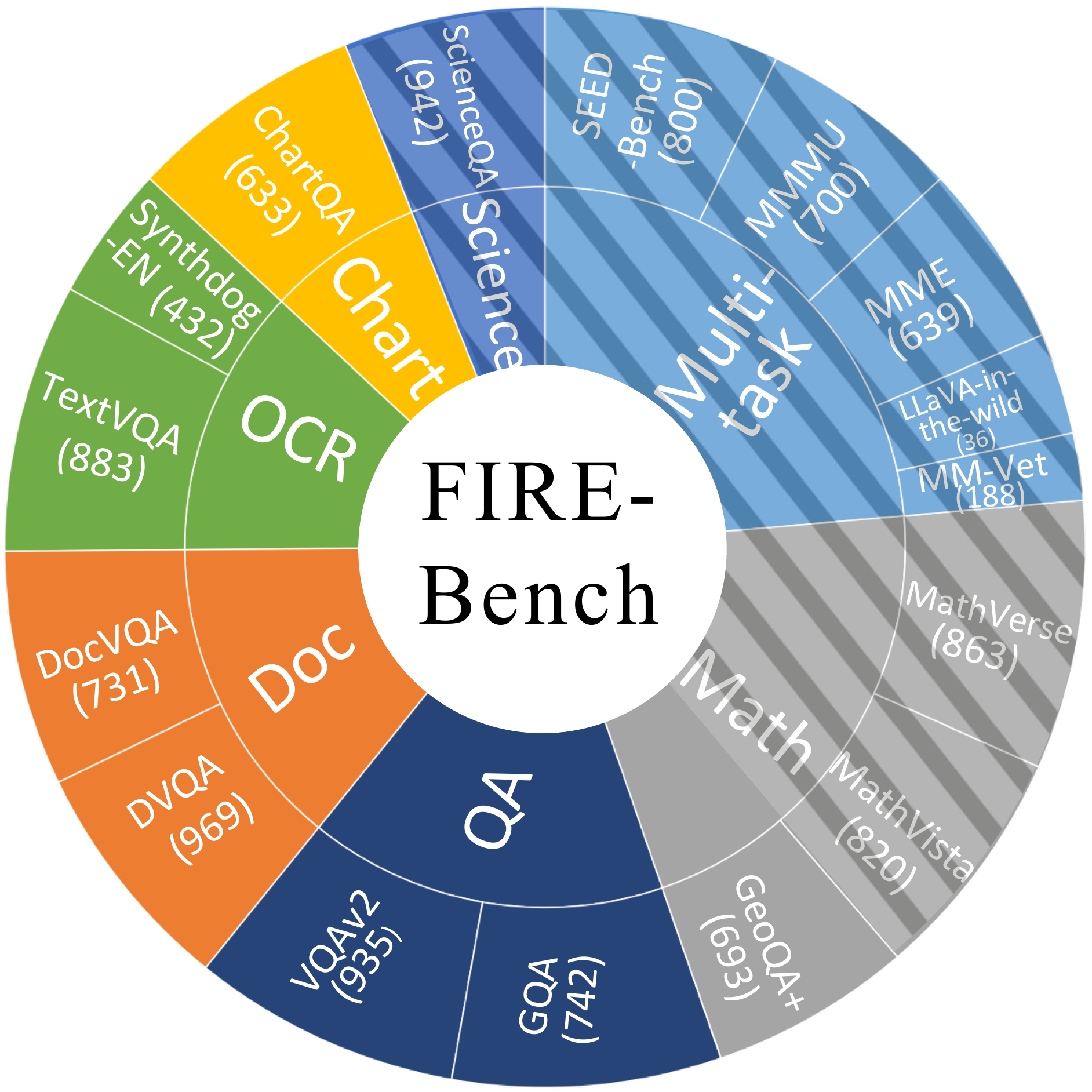}
  \label{fig:chart3_bench_pdf2ppt}}
    \caption{Data sources in FIRE. Shaded are new data sources in FIRE-Bench.
 }
\vspace{15pt}
	\label{fig:dataset_distribution}
\end{figure}


\section{\texorpdfstring{\acf{fire}}{}}
\vskip -0.1in

This section presents the \acs{fire} dataset, outlining its task definition, data collection methodology for FIRE-100K and FIRE-1M, and the creation of FIRE-Bench. Finally, we provide an analysis of FIRE.

\subsection{Task Definition}
\label{subsec:data_collect}
\vskip -0.1in

\noindent\textbf{Data Source.}
To enhance the diversity and comprehensiveness of our dataset, we compile more than 1.1M image-instruction-response triples from $27$ source datasets (more details can be found in \cref{appendix:data_source}), being used to generate FIRE-100K, FIRE-1M, and FIRE-Bench, as shown in~\cref{fig:dataset_distribution}.
These datasets cover tasks including visual question answering, image captioning, complex reasoning, OCR, chart/table/document analysis, math problems, science question answering \etc.

\noindent\textbf{Data format.}
We formulate our data as $ \{I, q, gt, \{r_i, f_i\}_{i=1}^n \}$, where $I$ denotes the image, $q$ is the instruction, $gt$ is the ground truth answer, and $\{r_i, f_i\}_{i=1}^n$ corresponds to the conversations in $n$ turns. 
In the $i$-th turn, $r_i$ is the response from VLMs, composed of the thought (how to refine the response based on feedback) and a new answer; $f_i$ is the feedback, involving a score $a_i$ (0-10) for the response $r_i$ and textual comments.





\subsection{\acs{fire}-100K}
\label{subsec:100kdata_generation}
\begin{figure}
    \centering
    \includegraphics[width=0.96\linewidth]{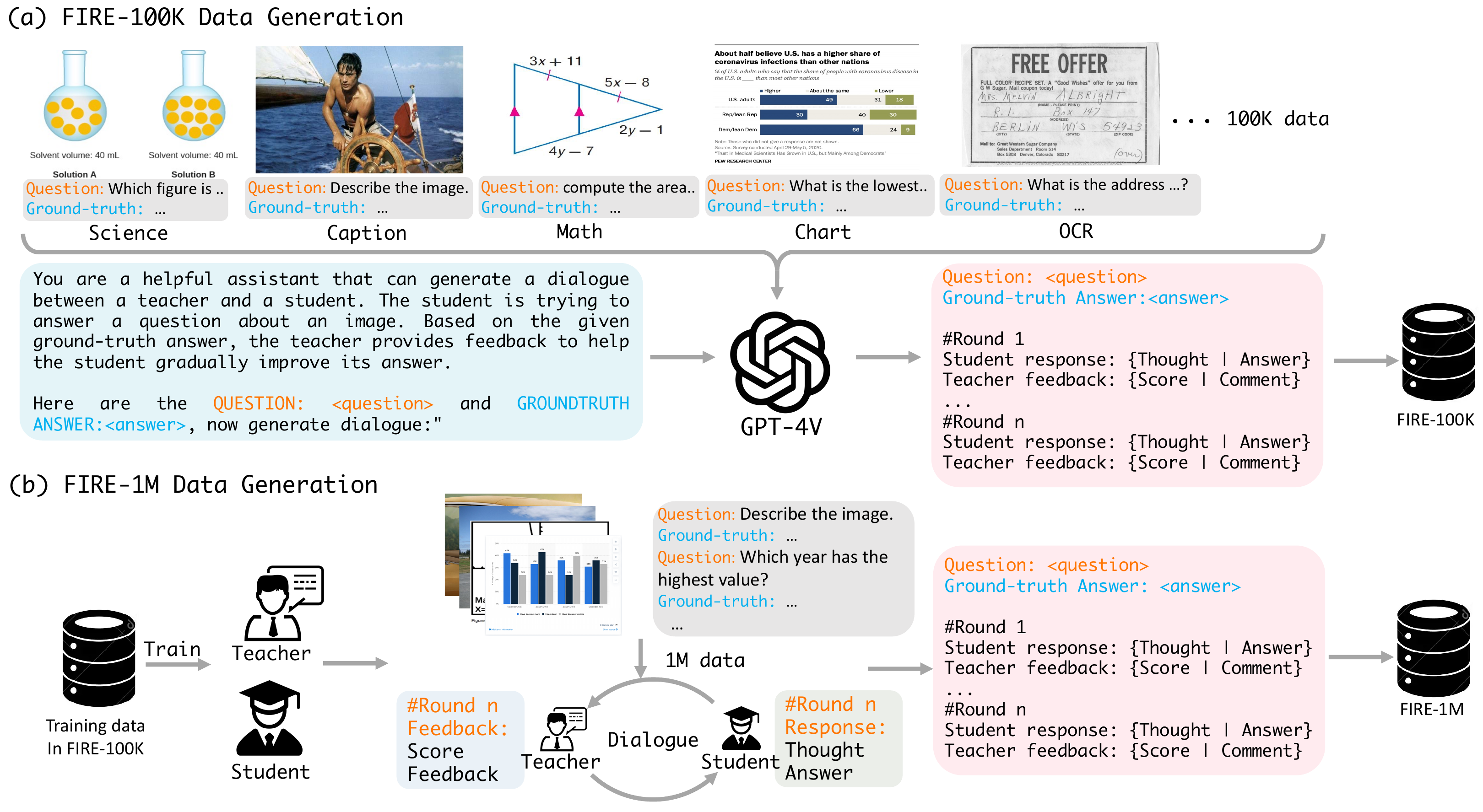}
    \caption{The pipeline to create FIRE-100K and FIRE-1M data.}
    \label{fig:data_gathering}
    \vskip -0.15in
\end{figure}

\vskip -0.1in
We feed images, instructions, ground truth answers from $18$ datasets, and a designed textual prompt to GPT-4V that generates high-quality feedback-refinement conversations in a one-go manner, as shown in~\cref{fig:data_gathering} (a).
We ask GPT-4V to play two roles: a student and a teacher, and generate a conversation between the two roles, where the student's responses are improved by incorporating feedback from the teacher. 
After generation, we filter out low-quality conversations with no score improvements or more than $6$ turns, since we expect that VLMs could learn to quickly and efficiently improve their responses from our data.
Finally, we obtain 100K conversations, shown in~\cref{fig:chart_100k_pdf2ppt}.


\subsection{\acs{fire}-1M}
\vskip -0.1in
\label{subsec:1mdata_generation}
We use FIRE-100K to fine-tune LLaVA-NeXT~\cite{llava1.6} and obtain two models: \studentmodel and \feedbackmodel, which are used to act as the student and the teacher, respectively (training details are shown in~\cref{sec:model}).
We sample 1M data from $18$ source datasets and generate feedback-refinement conversations via the following steps, as shown in~\cref{fig:data_gathering} (b).
(1) We feed an image and instruction to the student that generates a response. 
(2) We feed the image, instruction, ground truth answer, and the response to the teacher that generates feedback. If the score $a$ in the feedback $a\geq8$ or the number of turns exceeds $3$, we stop the conversation; otherwise, we go to step (3).
(3) We feed the feedback to the student that generates a refined response and go back to step (2).
Finally, we obtain 1M data, shown in~\cref{fig:chart_100k_pdf2ppt}

\subsection{\acs{fire}-Bench}
\vskip -0.1in
To comprehensively evaluate the feedback-refining ability of VLMs, we introduce FIRE-Bench, containing 11K high-quality feedback-refinement conversations. 
As shown in~\cref{fig:chart3_bench_pdf2ppt}, \acs{fire}-Bench is derived from $16$ source datasets, including $8$ seen datasets (test splits) from \acs{fire}-100K and \acs{fire}-1M, as well as $8$ new datasets from recently-proposed popular multimodal benchmarks, which is used to evaluate the generalization of the feedback-refining ability across different types of tasks.
Similar to \acs{fire}-100K, we sample 11K examples from the data sources and prompt GPT-4V to generate the feedback-refinement conversations.

\subsubsection{Evaluation Settings}
\begin{figure}
    \centering
\includegraphics[width=1\linewidth]{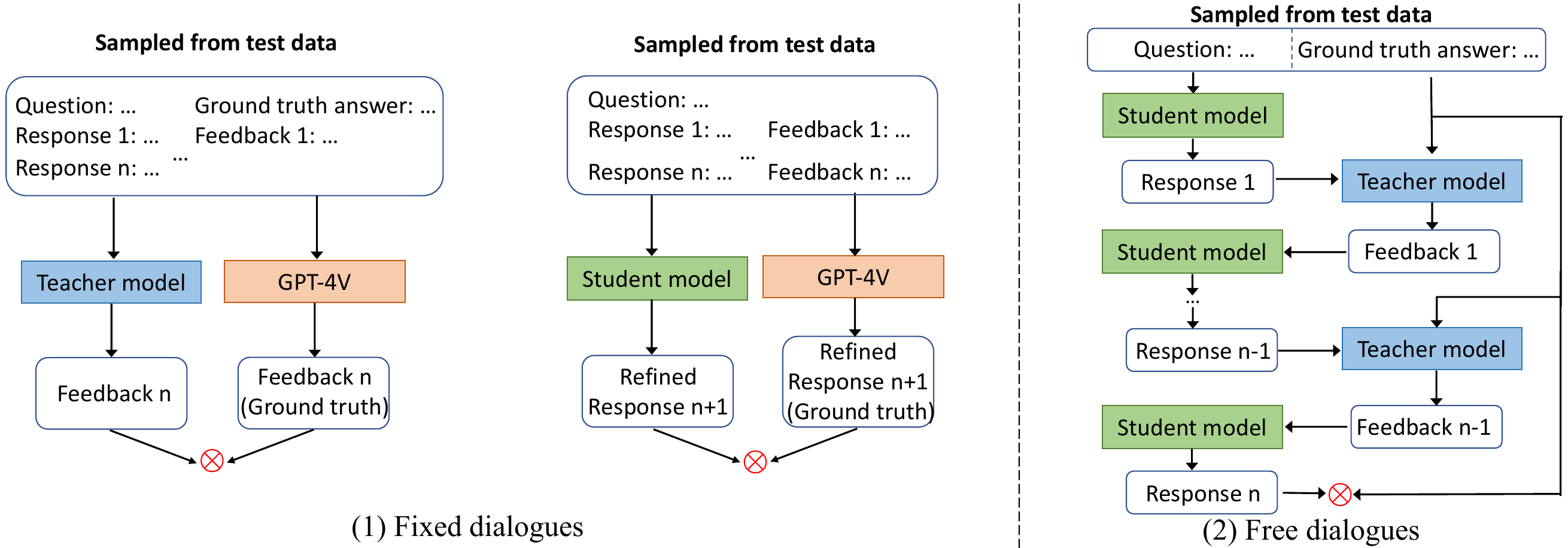}
    \caption{We use two settings to evaluate student and teacher models.}
    \label{fig:test_setting}
    \vskip -0.15in
\end{figure}

\vskip -0.1in
We design two evaluation settings: fixed dialogues and free dialogues to evaluate the performance of the student and teacher models, as shown in~\cref{fig:test_setting}.

\noindent\textbf{Fixed Dialogues.}
In fixed dialogues, we evaluate whether the student and teacher models can generate appropriate responses and feedback given the conversation history, 
and their performance is evaluated by being compared with GPT-4V generated feedback and response, using
the BLEU~\cite{papineni2002bleu} and CIDEr~\cite{vedantam2015cider} metrics to measure the textual alignment. 
For the predicted score $\hat{a_i}$ in feedback, 
we regard the score $a_i$ generated by GPT-4V as the ground truth and adopt \emph{mean absolute error (MAE)}: $MAE= \frac{1}{K} \sum_{k=1}^{K} |a_k-\hat{a_k}|$,
where there are $K$ test data totally. The teacher model may fail to follow instructions and does not generate a score in feedback for some cases. Here, we simply set $|a_i-\hat{a}_i|=10$ for these cases.

\noindent\textbf{Free Dialogues.}
We use a student model and a teacher model to perform free dialogues and evaluate how fast and how much the student model can improve its answers based on the feedback from the teacher model. 
The stopping condition for dialogues is that the obtained scores from the teacher model do not increase or exceed a pre-defined threshold (we set $8$ in experiments).

We introduce four metrics: average turn (AT), average dialogue refinement (ADR), average turn refinement (ATR), and refinement ratio (RR) for free dialogues.

(1) \emph{Average Turn (AT)}. The AT metric evaluates how fast a VLM could achieve a satisfactory result based on feedback. We measure the number of turns $n_k$ in the conversation to solve the $k$-th data, where VLMs refine their responses until the obtained score exceeds the pre-defined threshold. 
We set a punishment number as $p=10$, the maximum number of turns as $n_{max}=5$. If VLMs fail to obtain a satisfactory score in $n_{max}$ turns, then $n_k=p$. 
For clearer comparisons with the baseline model (\emph{e.g.}, the original LLaVA-NeXT model), we normalize it according to the AT of the baseline model, 
\vskip -0.2in
\begin{equation}
\begin{aligned}
AT= \frac{1}{K} \sum_{k=1}^{K} n_k / T_{baseline},
\end{aligned}
\end{equation}
\vskip -0.1in
where $T_{baseline}$ is the average turn of the baseline model. A smaller value of AT means better performance.

(2) \emph{Average Dialogue Refinement (ADR)}. The ADR metric evaluates how much knowledge VLMs could learn from feedback in a dialogue.
In solving the $k$-th data, we use $a_{k,1}$ to denote the obtained score for the initial response and use $a_{k,n_k}$ to denote the obtained score for the response in the final turn. ADR averages the score improvements of each conversation as
\vskip -0.2in
\begin{equation}
\begin{aligned}
ADR= \frac{1}{K} \sum_{k=1}^{K}  a_{k,n_k}-a_{k,1}.
\end{aligned}
\end{equation}
\vskip -0.1in
A larger value of ADR means better performance.

(3) \emph{Average Turn Refinement (ATR)}. ATR evaluates how much knowledge VLMs could learn from feedback in one turn.
ATR averages the score improvements in each turn of $K$ samples as
\vskip -0.2in
\begin{equation}
\begin{aligned}
ATR= \frac{1}{K} \sum_{k=1}^{K}  \frac{1}{n_k-1} (a_{k,n_k}-a_{k,1}).
\end{aligned}
\end{equation}
\vskip -0.1in
A larger value of ATR means better performance.

(4) \emph{Refinement Ratio (RR)}. RR measures the proportion of data that have a wrong initial response and a correct final response (\emph{i.e.}, how much data are corrected based on feedback), computed by
\vskip -0.2in
\begin{equation}
\begin{aligned}
RR= \frac{1}{K} \sum_{k=1}^{K}  \mathbbm{1}_{a_{k,n_k} \geq 8} - \mathbbm{1}_{a_{k,1} \geq 8},
\end{aligned}
\end{equation}
\vskip -0.1in
where $\mathbbm{1}_{a_{k,n_k} \geq 8}$ means if $a_{k,n_k} \geq 8$, $\mathbbm{1}_{a_{k,n_k} \geq 8}=1$, and $0$ otherwise. A larger value of RR means better performance. 
Note that, for the $k$-th sample, if $n_k=1$, we remove it from the K samples to compute AT, ADR, ATR, and RR.




\subsection{Dataset Analysis} \label{subsec:data_analysis}
\begin{figure}[tb]
  \centering
  \subfigure[Score in the first turn]{
  \includegraphics[width=0.31\textwidth]{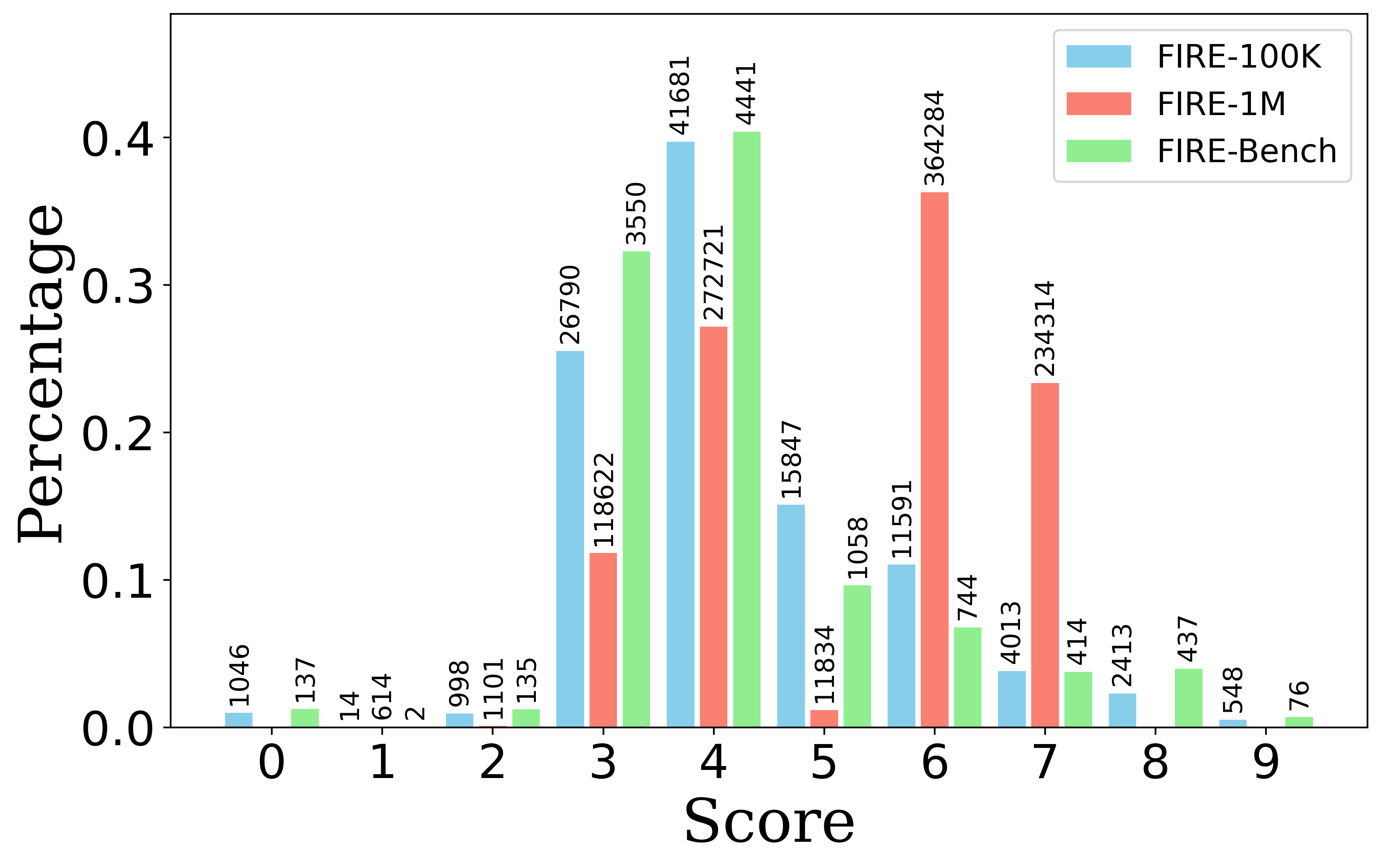}\label{fig:first_round_score}}
  \subfigure[Improved score per turn]{
  \includegraphics[width=0.31\textwidth]{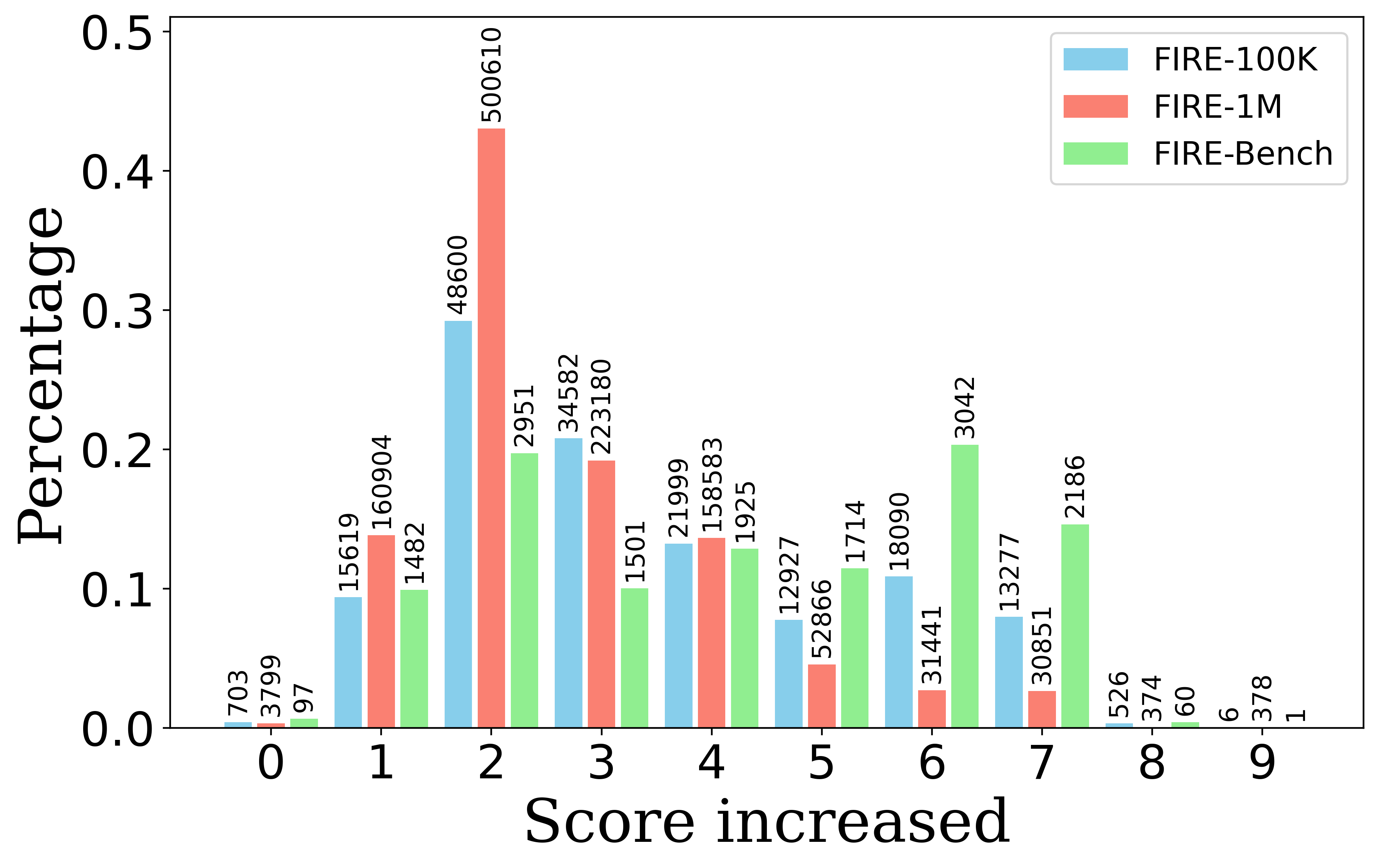}\label{fig:improved_score_round}}
  \subfigure[Improved score per dialogue]{
  \includegraphics[width=0.31\textwidth]{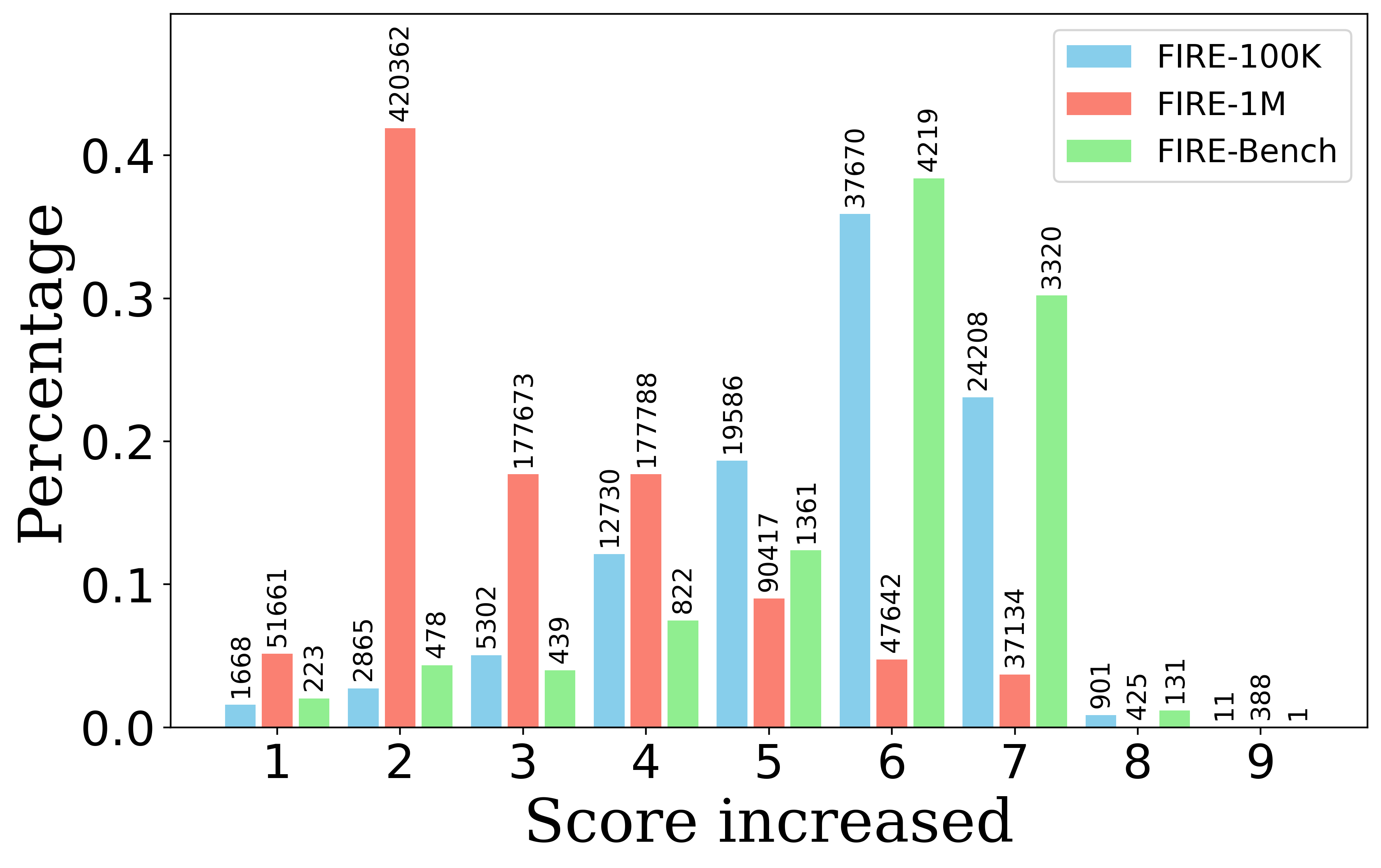}\label{fig:improved_score_dialogue}}
  \subfigure[Turn number]{
  \includegraphics[width=0.31\textwidth]{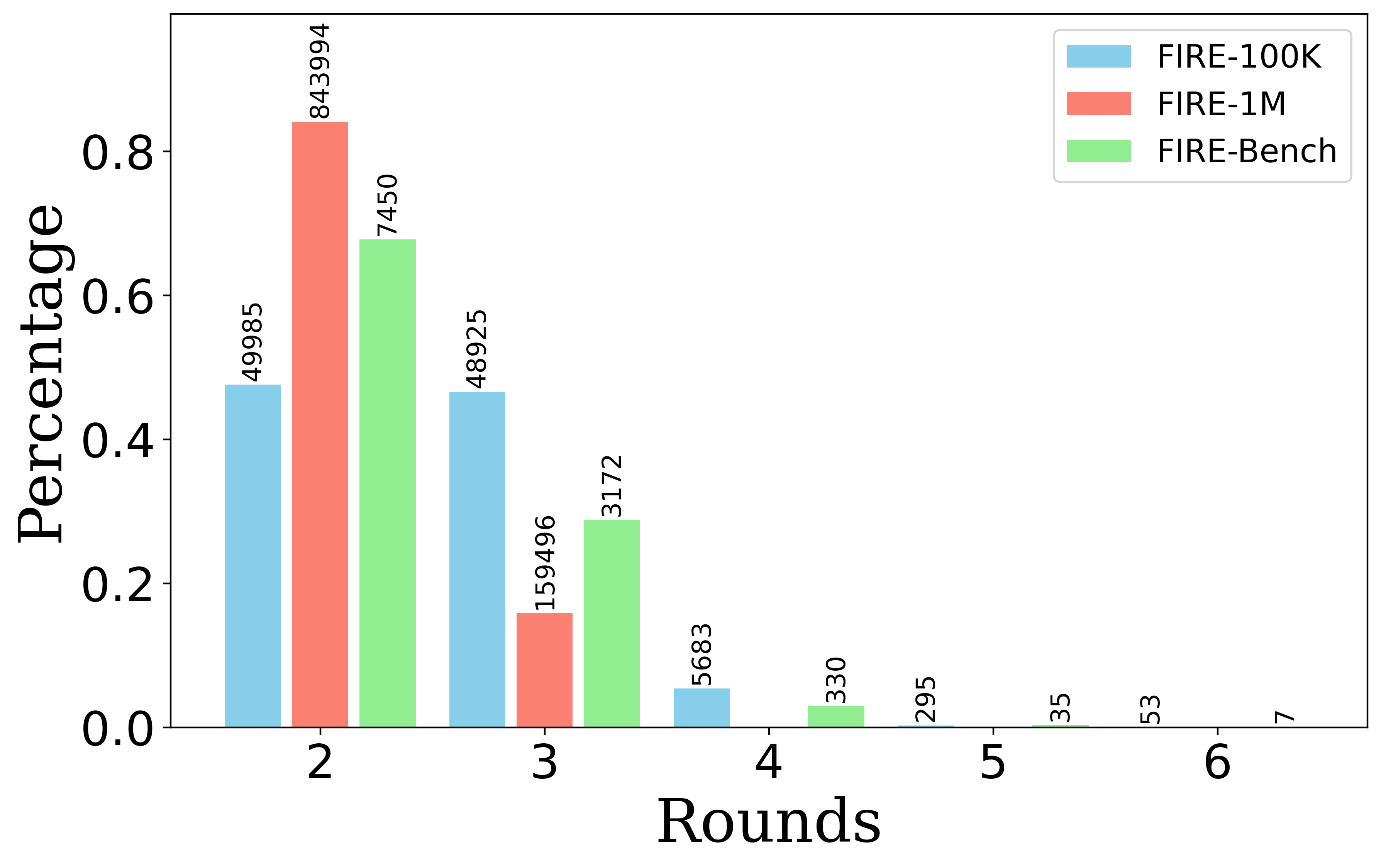}\label{fig:round}}    
  \subfigure[Length of response]{ 
  \includegraphics[width=0.31\textwidth]{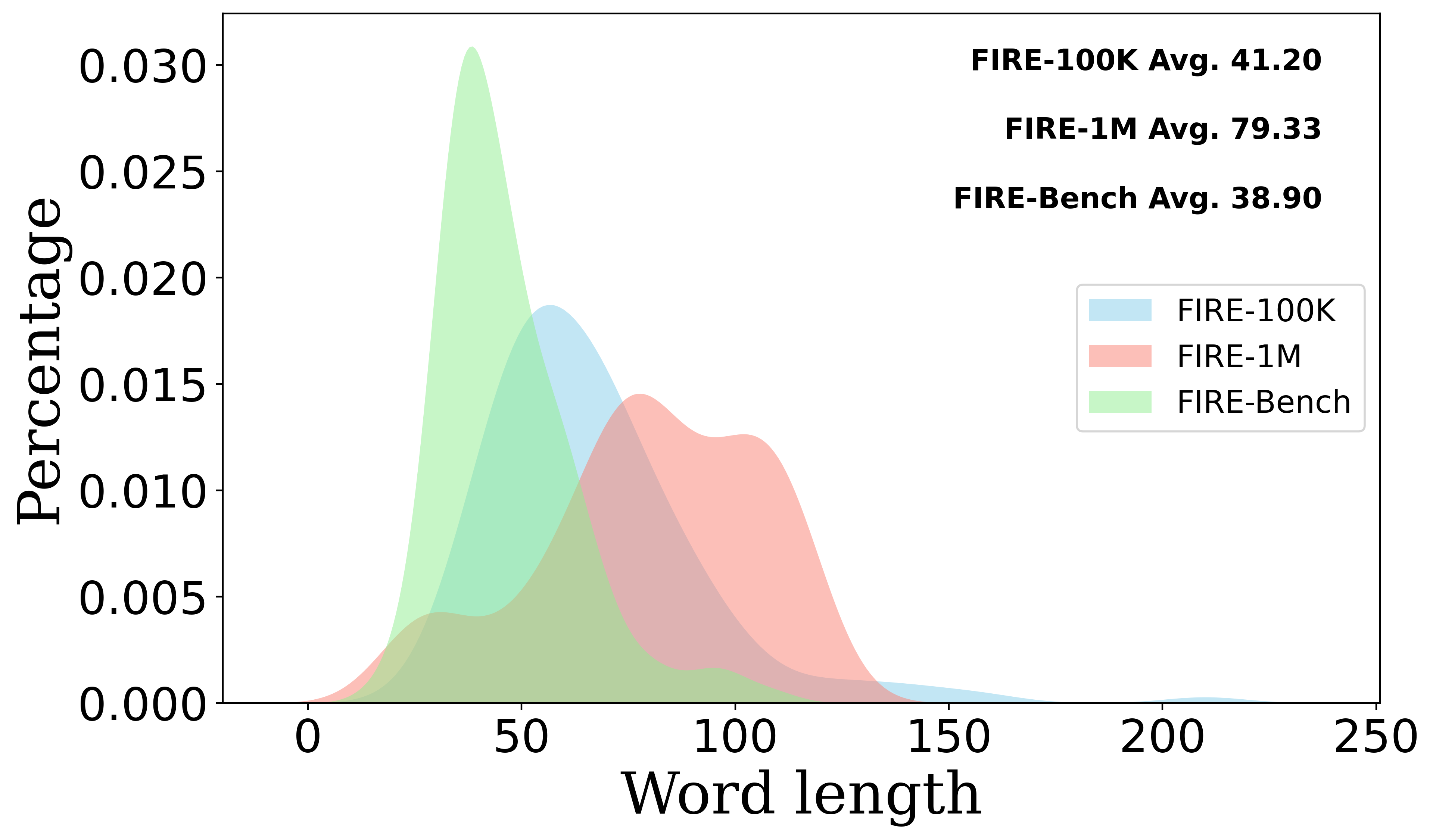}\label{fig:student_word_length}} 
  \subfigure[Length of feedback]{
  \includegraphics[width=0.31\textwidth]{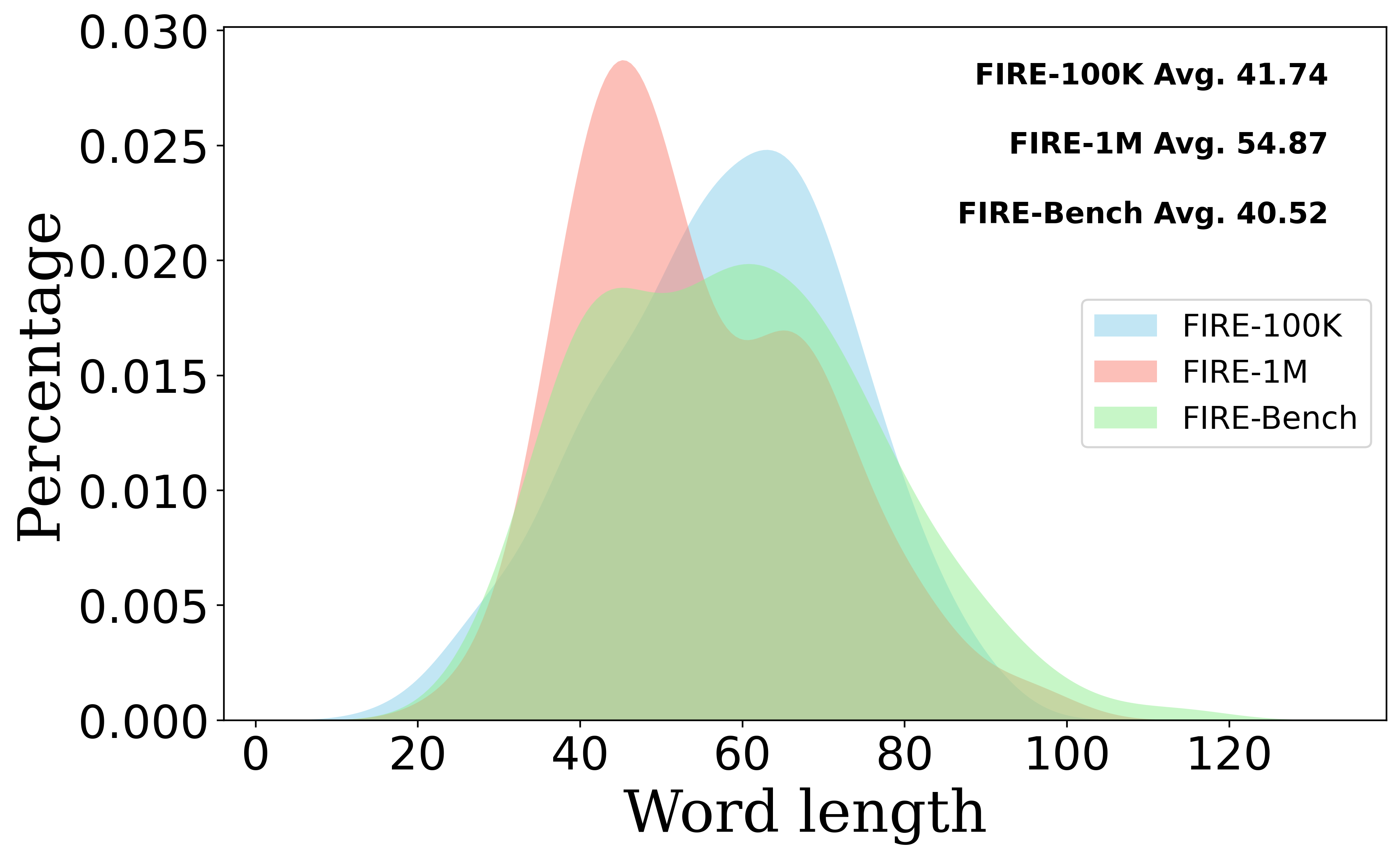}\label{fig:teacher_word_length}} 
  \vskip -0.1in
  \caption{Data statistics on \acs{fire}-100K, \acs{fire}-1M, \acs{fire}-Bench.}
  \label{fig:statistics}
  \vskip -0.05in
\end{figure}

\vskip -0.1in
We provide three key statistics: score, turn, and length, for the collected feedback-refinement conversations.
\noindent\textbf{Score.} 
We show the distribution of initial scores in~\cref{fig:first_round_score}, which reflects the starting state of the conversation. They mainly fall in the interval $[3,8]$, showing that FIRE covers diverse starting states of conversations.
Improved scores per turn are shown in~\cref{fig:improved_score_round}, which reflects the learning effect. It ranges from $[2,8]$, similar to actual situations, where high improvements are obtained in easy cases and small improvements are obtained in hard cases, showcasing the diversity of data.
Improved scores per dialogue are shown in~\cref{fig:improved_score_dialogue}, and the improvements in most cases are 5-7, demonstrating the data quality of FIRE, where most data have obvious improvements, helping VLMs to efficiently learn to improve their responses.
The score distributions of FIRE-100K, FIRE-1M, and FIRE Bench are not completely consistent, making the data more diverse.
\noindent\textbf{Turn.} The turn distribution of conversations is shown in~\cref{fig:round}. Most conversations have 2-4 turns, indicating an efficient and concise feedback process. This measure suggests that most conversations reach a satisfactory level of refinements. 
A small number of turns in FIRE informs VLMs to perform effective dialogues.
\noindent\textbf{Length.} The length distributions of responses and feedback are shown in~\cref{fig:student_word_length} and~\cref{fig:teacher_word_length}, respectively. Most responses or feedback are less than $100$ words. It shows concise dialogues in FIRE, aligning with real-world scenarios where users typically engage in brief exchanges rather than lengthy discussions.

\section{Model}
\label{sec:model}
\vskip -0.1in
Our model architecture has the same design as LLaVA-NeXT-8B~\cite{llava1.5} that uses CLIP~\cite{clip} as a frozen image encoder with a two-layer multi-layer perceptron vision-language connector. For the LLM part, we use the same architecture as the LLaMA3-8B~\cite{llama3}.
We use LLaVA-NeXT-8B to initialize the VLMs and use LoRA to fine-tune the LLaVA-NeXT-8B for a student model and a teacher model.


\subsection{Student Model} 
\vskip -0.1in
Given an $n$-turn conversation $\{I, q, gt, \{r_i, f_i\}_{i=1}^n \}$, we train a student model to fit responses $r_i$ for $i\geq2$ using the cross-entropy loss,
\vskip -0.2in
\begin{equation}
\begin{aligned}
\min \mathbb{E}_{(I, q, gt, \{r_i, f_i\}_{i=1}^n )\sim \mathbb{D} }\left[ - \sum_{i=2}^{n} \log P(r_i |I, q, \{r_j, f_j\}_{j=1}^{j=i-1} )  \right],
\end{aligned}
\end{equation}
\vskip -0.05in
where $\mathbb{D}$ is the used dataset. We first use FIRE-100K as $\mathbb{D}$ to train a student model \studentmodel, then use all training data (FIRE-100K and FIRE-1M) to train a final student model \model.

\subsection{Teacher Model}
\vskip -0.1in
Given a $n$-turn conversation $\{I, q, gt,  \{r_i, f_i\}_{i=1}^n \}$, we train a teacher model to fit the feedback $f_i$ for $i\geq1$ using the cross-entropy loss,
\vskip -0.2in
\begin{equation}
\begin{aligned}
\min \mathbb{E}_{(I, q, gt, \{r_i, f_i\}_{i=1}^n ) \sim \mathbb{D} }\left[ - \sum_{i=1}^{n} \log P(f_i |I, q, gt, \{r_j, f_j\}_{j=1}^{j=i-1},r_i )  \right],
\end{aligned}
\end{equation}
\vskip -0.05in
where we use FIRE-100K as $\mathbb{D}$ and obtain the teacher model \feedbackmodel.


\begin{table}
\caption{Comparisons between LLaVA-NeXT-8B and \studentmodel on 10 benchmarks.
Benchmark names are abbreviated for space limits. GQA~\cite{hudson2019gqa}; VQAv2~\cite{goyal2017making};VizWiz~\cite{gurari2018vizwiz};  TextVQA~\cite{singh2019towards}; SQA$^I$:ScienceQA-IMG~\cite{lu2022learn}; LLaVA$^W$: LLaVA-Bench-in-the-wild~\cite{llava};MMB: MMBench~\cite{liu2023mmbench}; MME$^P$: MME Perception~\cite{fu2024mme}; MME$^C$: MME Cognition~\cite{fu2024mme}; MM-Vet~\cite{yu2023mm}.
}
\vskip -0.08in
\label{table:accuracy}
\centering
\footnotesize
\resizebox{0.9\columnwidth}{!}{
\setlength{\tabcolsep}{1.5pt}
\begin{tabular}{ c | c c c c c c c c c c }
    \toprule
   Method & GQA &	VQAv2 &	VizWiz & TextVQA &	SQA$^I$ & LLaVA$^W$  & MMB  & MME$^P$& MME$^C$  & MM-Vet \\
   \hline
   LLaVA-NeXT-8B & \textbf{65.9} & 79.0 & 52.0 & \textbf{69.8} & \textbf{77.3} & 78.5 & 74.4 & \textbf{1546.0} & \textbf{331.4} & 44.9\\
   \model & 65.8 & \textbf{82.9} & \textbf{59.8} & 68.4 & 76.8 & \textbf{81.5}  & \textbf{78.5} & 1534.8 & 321.1 & \textbf{45.3}\\  
    \bottomrule
\end{tabular}
}
\end{table}

\section{Experiments}
\vskip -0.15in
\label{sec:experiment}
We conduct experiments to evaluate both the student and teacher models trained on FIRE. We first provide experimental details and then comprehensively evaluate models in multiple settings.

\begin{table}
\caption{Results of the student model in fixed dialogues.}
\label{table:fixed_dialogue_vlms}
\centering
\footnotesize
\vskip -0.05in
\begin{tabular}{ c | c c c c c }
    \toprule
   Model & BLEU-1 ($\uparrow$) & BLEU-2 ($\uparrow$) & BLEU-3 ($\uparrow$) & BLEU-4 ($\uparrow$) & CIDEr ($\uparrow$) \\
   \hline
   LLaVA-NeXT-8B & 0.33 & 0.23 & 0.17 & 0.13 & 0.60 \\
   \model & \textbf{0.54} & \textbf{0.46} &\textbf{ 0.39} & \textbf{0.34} & \textbf{2.36} \\
    \bottomrule
\end{tabular}
  \vskip -0.1in
\end{table}

\begin{table}
\caption{Results of the teacher model in fixed dialogues.}
\label{table:fixed_dialogue_feedback}
\centering
\footnotesize
\vskip -0.1in
\resizebox{1\columnwidth}{!}{
\begin{tabular}{ c | c c c c c c }
    \toprule
   Model  & BLEU-1 ($\uparrow$) & BLEU-2 ($\uparrow$) & BLEU-3 ($\uparrow$) & BLEU-4 ($\uparrow$) & CIDEr ($\uparrow$) & MAE ($\downarrow$)\\
   \hline
   LLaVA-NeXT-8B & 0.34  & 0.21 & 0.15 & 0.10 & 0.51 &	1.88 \\
   \feedbackmodel & \textbf{0.55}  & \textbf{0.45} & \textbf{0.39} & \textbf{0.33} & \textbf{2.27} & \textbf{0.30}\\
    \bottomrule
\end{tabular}
}
  \vskip -0.05in
\end{table}

\subsection{Experimental Details}
\vskip -0.1in
\noindent\textbf{Training Data.} 
To avoid the catastrophic forgetting issue, we combine the training data in FIRE with the LLaVA-665K~\cite{llava} (released by Open-LLaVA-1M~\cite{openllava1m}) to train the student and teacher models. 



\noindent
\textbf{Training Details.}
In the training process of both the student and teacher models, we freeze the image encoder and the image-language connector, and fine-tune the language decoder using LoRA~\cite{hu2021lora}. In the implementation of LoRA, we set the rank as $64$ and only apply LoRA on the query and key projection matrices in all attention layers of the language decoder. This setting only involves $0.4\%$ parameters of LLaMA3-8B. We use the AdamW optimizer, where a cosine annealing scheduler is employed, the learning rate is $2e-4$, the batch size is $64$, and we train $1$ epoch over all data.
The training process for a student (or teacher) model requires about 128 A100-80GB GPU hours.

\vskip -0.15in
\subsection{Evaluation in Instruction Following}
\vskip -0.1in
Considering that fine-tuning VLMs may encounter the catastrophic forgetting problem, we evaluate the instruction-following ability of \model, using $10$ commonly used multimodal benchmarks, as shown in~\cref{table:accuracy}. Our model achieves comparable performance to the original LLaVA-NeXT-8B model, showing that we do not compromise the instruction-following ability when learning the feedback-refining ability.


\subsection{Evaluation in Fixed Dialogues}
\vskip -0.1in
We evaluate the performance of \model, and \feedbackmodel in fixed dialogues.  
The evaluation of \model is shown in~\cref{table:fixed_dialogue_vlms}, where we report the results of BLEU-1, BLEU-2, BLEU-3, BLEU-4, and CIDEr. 
The performance of \feedbackmodel is shown in~\cref{table:fixed_dialogue_feedback}, where we report the results of BLEU-1, BLEU-2, BLEU-3, BLEU-4, CIDEr, and MAE.
We observe that using FIRE, \model and \feedbackmodel generates good responses and feedback, having better performance than the original LLaVA-NeXT-8B model on all metrics.
\model could well refine the responses, like GPT-4V. 
\feedbackmodel can generate more accurate feedback, including comments (see BLEU and CIDEr) and scores (see MAE), 
demonstrating the effectiveness of our teacher model \feedbackmodel that can discover undesirable responses.


\begin{table}
\caption{Results in free dialogues overall test data in FIRE.}
\vskip -0.1in
\label{table:test_all}
\centering
\footnotesize
\begin{tabular}{ c |  c c c c }
    \toprule
   Model & AT ($\downarrow$) & ADR ($\uparrow$) & ATR ($\uparrow$) & RR ($\uparrow$) \\
   \hline
  LLaVA-NeXT-8B  & 1 &	0.97&	0.41&	0.25 \\
   \studentmodel-8B  &  0.92&	1.27&	0.55&	0.34\\   
   \model-8B  &\textbf{ 0.84} &	\textbf{1.56}&	\textbf{0.66}&	\textbf{0.39} \\
    \bottomrule
\end{tabular}
\end{table}


\begin{table}
\caption{Detailed test results (AT ($\downarrow$), ADR ($\uparrow$), ATR ($\uparrow$), and RR ($\uparrow$)) on 8 seen source datasets.}
\vskip -0.05in
\label{table:test_seen}
\centering
\tiny
\resizebox{1\columnwidth}{!}{
\setlength{\tabcolsep}{1pt}
\begin{tabular}{ c|c c c c|c c c c|c c c c|c c c c}
    \bottomrule
   \multirow{2}{*}{Model}  & \multicolumn{4}{c|}{VQAv2} & \multicolumn{4}{c|}{GQA} & \multicolumn{4}{c|}{TextVQA} & \multicolumn{4}{c}{ChartQA} \\
    \cline{2-17}
    & AT & ADR & ATR & RR & AT & ADR & ATR & RR & AT & ADR & ATR & RR & AT & ADR & ATR & RR \\   
    \hline
    LLaVA-NeXT      & 1.00&	1.45&	0.42&	0.40    & 1.00&	1.51&	0.51&	0.43    & 1.00&	0.91&	0.34&	0.26   & 1.00&	0.71&	0.39&	0.25 \\
    \studentmodel   & 0.86&	1.83&	0.55&	0.54	& 0.81&	1.93&	0.63&	0.58	& 0.95&	1.20&	0.49&	0.33   & 1.07&	1.03&	\textbf{0.56}&	0.27 \\   
    \model          & \textbf{0.78}&	\textbf{2.08}&	\textbf{0.59}&	\textbf{0.56}	& \textbf{0.81}&	\textbf{2.06}&	\textbf{0.70}&	\textbf{0.58}	& \textbf{0.77}&	\textbf{1.51}&	\textbf{0.56}&	\textbf{0.42}   & \textbf{0.79}&	\textbf{1.15}&	0.53&	\textbf{0.36}\\
    \bottomrule
   \multirow{2}{*}{Model}  & \multicolumn{4}{c|}{DocVQA} & \multicolumn{4}{c|}{DVQA} & \multicolumn{4}{c|}{GEOQA+} & \multicolumn{4}{c}{Synthdog} \\
    \cline{2-17}
    & AT & ADR & ATR & RR & AT & ADR & ATR & RR & AT & ADR & ATR & RR & AT & ADR & ATR & RR \\    
    \hline
    LLaVA-NeXT      & 1.00&	0.97&	0.56&	0.24    & 1.00&	1.66&	\textbf{0.50}&	0.42    & 1.00&	0.14&	0.07&	0.08    & 1.00&	0.14&	0.05&	0.04\\
    \studentmodel   & 1.06&	0.84&	0.51&	0.22	& 0.79&	1.87&	0.46&	\textbf{0.51}	& \textbf{0.84}&	0.70&	0.33&	\textbf{0.28}	& \textbf{0.93}&	0.18&	0.07&	\textbf{0.08} \\   
    \model          & \textbf{0.81}&	\textbf{1.65}&	\textbf{0.97}&	\textbf{0.41}	& \textbf{0.74}&	\textbf{1.97}&	0.46&	0.50	& \textbf{0.84}&	\textbf{0.74}&	\textbf{0.35}&	0.27	& 0.95&	\textbf{0.19}&	\textbf{0.08}&	0.06\\
    \bottomrule
\end{tabular}
}
\vskip -0.05in
\end{table}

\begin{table}
\caption{Detailed test results (AT ($\downarrow$), ADR ($\uparrow$), ATR ($\uparrow$), and RR ($\uparrow$)) on 8 new source datasets.}
\vskip -0.05in
\label{table:test_unseen}
\centering
\tiny
\resizebox{1\columnwidth}{!}{
\setlength{\tabcolsep}{1pt}
\begin{tabular}{ c|c c c c|c c c c|c c c c|c c c c}
    \bottomrule
   \multirow{2}{*}{Model}  & \multicolumn{4}{c|}{MathVista} & \multicolumn{4}{c|}{MathVerse} & \multicolumn{4}{c|}{MMMU} & \multicolumn{4}{c}{MME} \\
    \cline{2-17}
    & AT & ADR & ATR & RR & AT & ADR & ATR & RR & AT & ADR & ATR & RR & AT & ADR & ATR & RR \\   
    \hline
    LLaVA-NeXT       & 1.00&	0.84&	0.45&	0.19           & 1.00&	0.14&	0.13&	0.08        & 1.00&	0.94&	0.53&	0.22       & 1.00&	1.31&	0.31&	0.21\\
    \studentmodel   & 0.89&	1.09&	0.68&	0.29	           & 0.95&	0.34&	0.30&	0.16	    & 0.86&	1.38&	0.81&	0.38  	   & \textbf{0.95}&	\textbf{2.20}&	\textbf{0.60}&	\textbf{0.39} \\   
    \model          & \textbf{0.83}&	\textbf{1.36}&	\textbf{0.77}&	\textbf{0.34}	          & \textbf{0.93}&	\textbf{0.65}&	\textbf{0.49}&	\textbf{0.17}	    & \textbf{0.80}&	\textbf{1.73}&	\textbf{1.05}&	\textbf{0.41}	   & 0.96&	2.04&	0.57&	0.36\\
    \bottomrule
   \multirow{2}{*}{Model}  & \multicolumn{4}{c|}{MM-Vet} & \multicolumn{4}{c|}{SEED-Bench} & \multicolumn{4}{c|}{ScienceQA} & \multicolumn{4}{c}{LLaVA-wild} \\
    \cline{2-17}
    & AT & ADR & ATR & RR & AT & ADR & ATR & RR & AT & ADR & ATR & RR & AT & ADR & ATR & RR \\    
    \hline
LLaVA-NeXT          & 1.00&	0.80&	0.31&	0.13    & 1.00&	2.30&	0.56&	0.48    & 1.00&	2.81&	0.70&	0.56    & 1.00&	0.45&	0.19&	0.03\\
    \studentmodel   & 0.97&	0.99&	0.48&	0.23	& 0.83&	3.18&	0.75&	0.68	& 0.98&	2.95&	0.78&	0.62	& 0.99&	0.79&	0.33&	\textbf{0.12 }\\   
    \model          & \textbf{0.87}&	\textbf{1.18}&	\textbf{0.60}&	\textbf{0.26}	& \textbf{0.81}&	\textbf{3.34}&	\textbf{0.84}&	\textbf{0.69}    &\textbf{ 0.83}&	\textbf{3.94}&	\textbf{1.08}&	\textbf{0.78}	& \textbf{0.96}&	\textbf{0.85}&	\textbf{0.50}&	\textbf{0.12} \\
    \bottomrule
\end{tabular}
}
\vskip -0.2in
\end{table}

\subsection{Evaluation in Free Dialogues}
\vskip -0.15in
We employ a student model and a teacher model to perform free dialogues.
We evaluate LLaVA-NeXT-8B, \studentmodel, and \model as the student model, and use \feedbackmodel to act as the teacher model.
We report the average turn (AT), average dialogue refinement (ADR), average turn refinement (ATR), and refinement ratio (RR) on FIRE-Bench.
Results are shown in~\cref{table:test_all}.
We observe that a LLaVA model trained on FIRE has improved feedback-refining ability. On the ADR, ATR, and RR metrics, \model achieves more than $50 \%$ improvements by LLaVA-NeXT, making an efficient user-agent interaction. 
Meanwhile, adding FIRE-1M to training data has better performance than only using FIRE-100K, showing the data quality of FIRE-1M.


We also show the detailed results on $8$ seen source datasets and $8$ new source datasets, as shown in~\cref{table:test_seen} and~\cref{table:test_unseen}, respectively.
Our models achieve improvements on both seen and new datasets, showing the generalization of feedback-refining ability across different types of data and tasks.

\begin{figure}[htbp]
\centering
    \subfigure[AT($\downarrow$)]{
  \includegraphics[width=0.47\textwidth]{{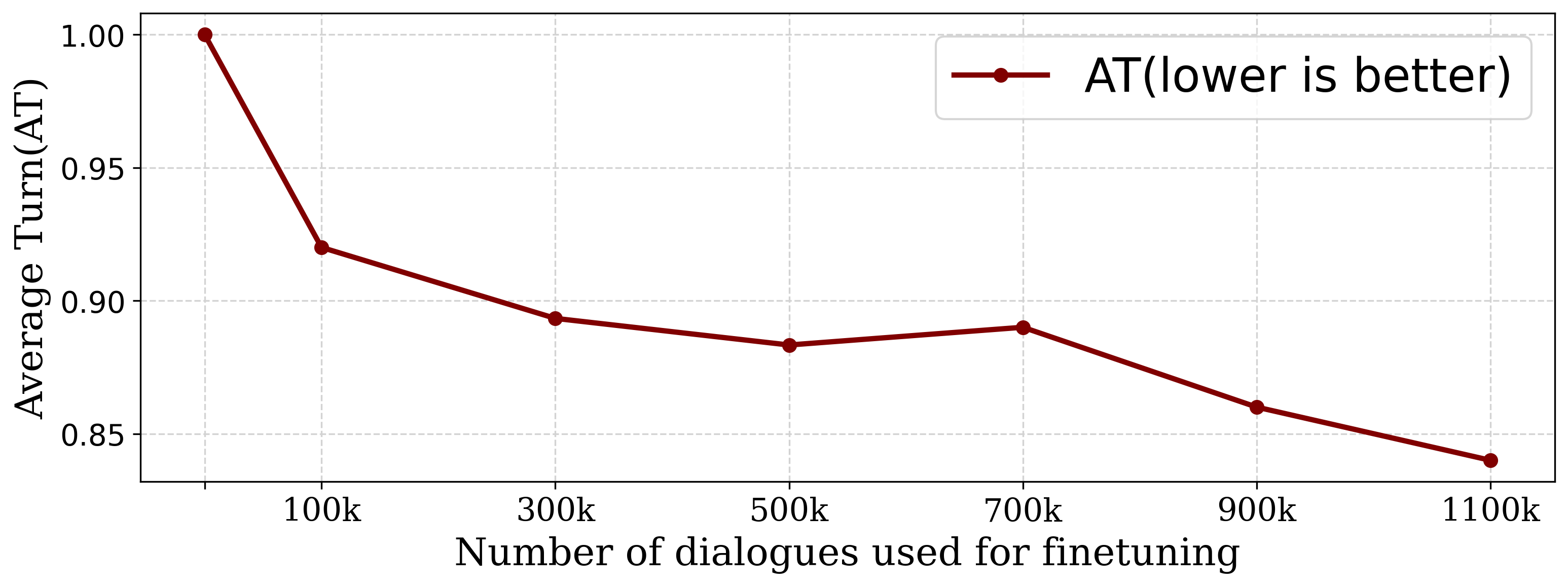}}}
    \subfigure[ADR($\uparrow$)]{
  \includegraphics[width=0.47\textwidth]{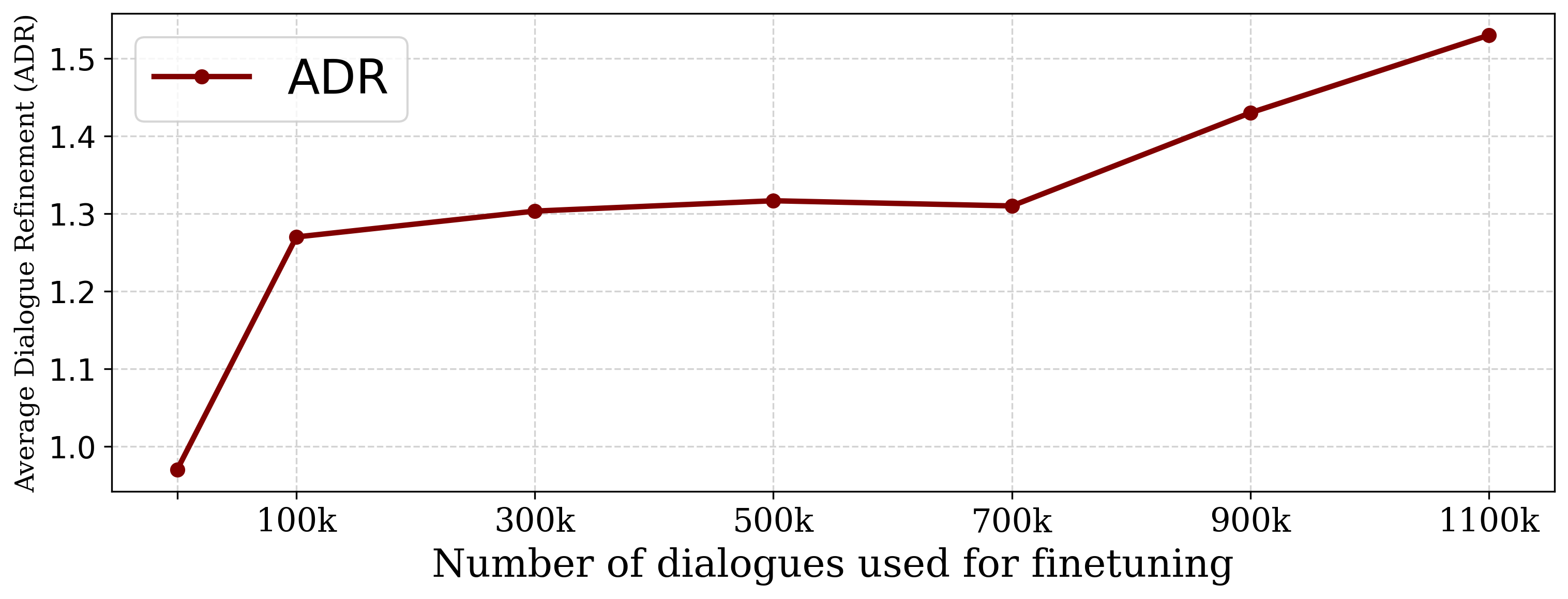}}
    \subfigure[ATR($\uparrow$)]{
  \includegraphics[width=0.47\textwidth]{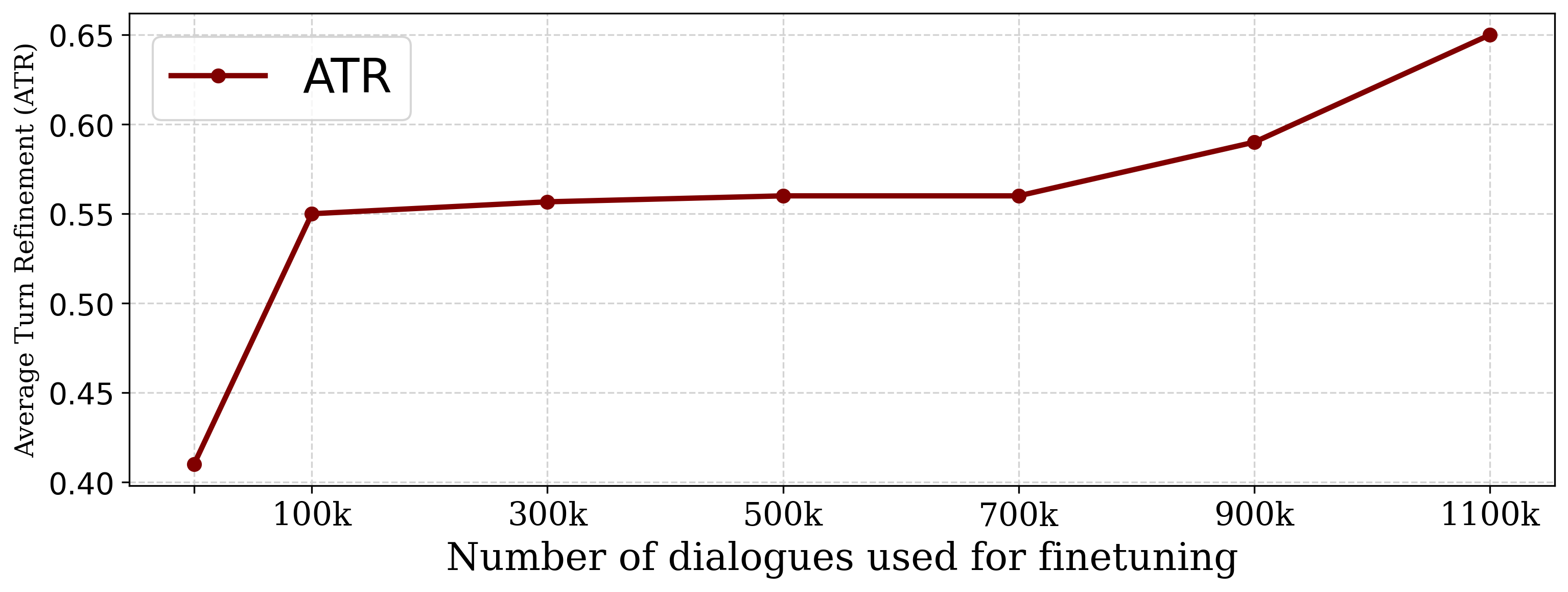}}
    \subfigure[RR($\uparrow$)]{
  \includegraphics[width=0.47\textwidth]{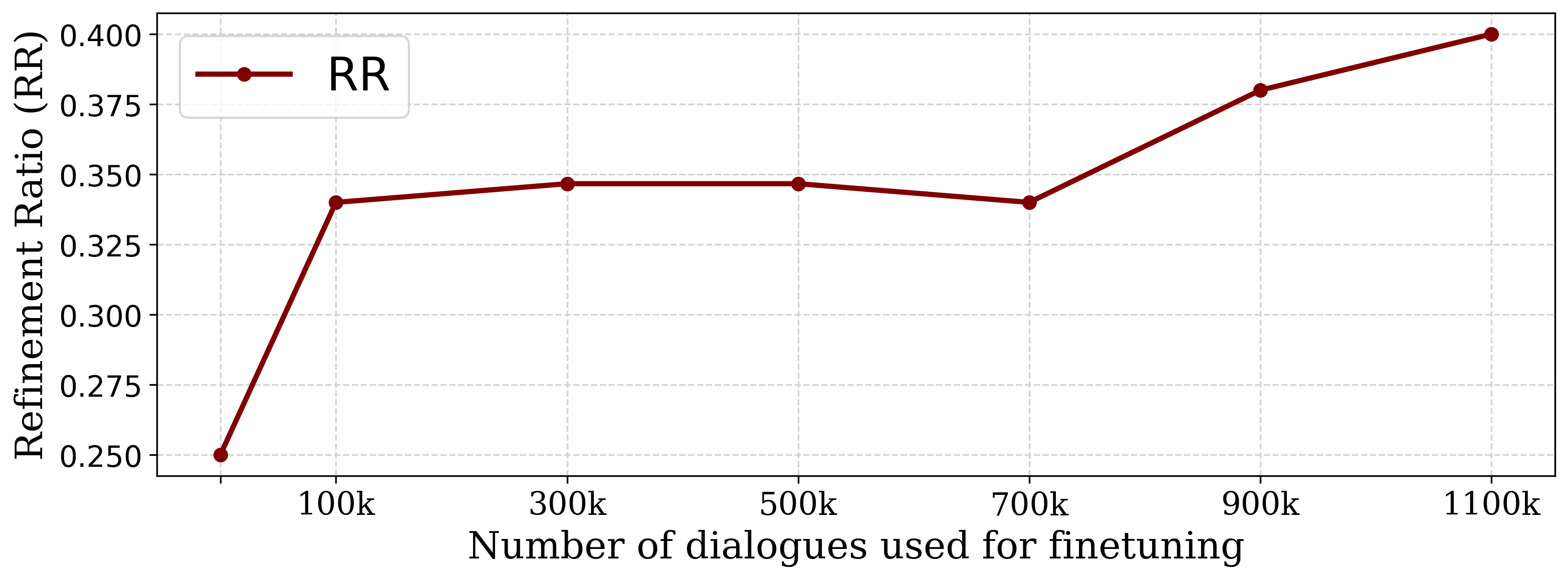}}
  \caption{Impact of training set size on model performance.}
  \label{fig:supp-score-number}
\end{figure}

\subsection{Ablation Studies}
In~\cref{fig:supp-score-number}, we evaluate the feedback-refining ability of VLMs using different amounts of training data from the FIRE dataset. Concretely, we first use the FIRE-100K data. Then, we gradually sample data from FIRE-1M, varying from 200K to 1000K, combined with FIRE-100K to train the LLaVA-NEXT-8B model. 
Overall, the results indicate that more training data leads to better performance across all evaluated metrics. The substantial improvements, particularly with the initial 100K dialogues and the noted enhancement at around 700K dialogues, demonstrate the high quality of the FIRE dataset and the model's emergent capabilities with more training data.


 
\begin{figure}
    \centering
    \includegraphics[width=0.5\textwidth]{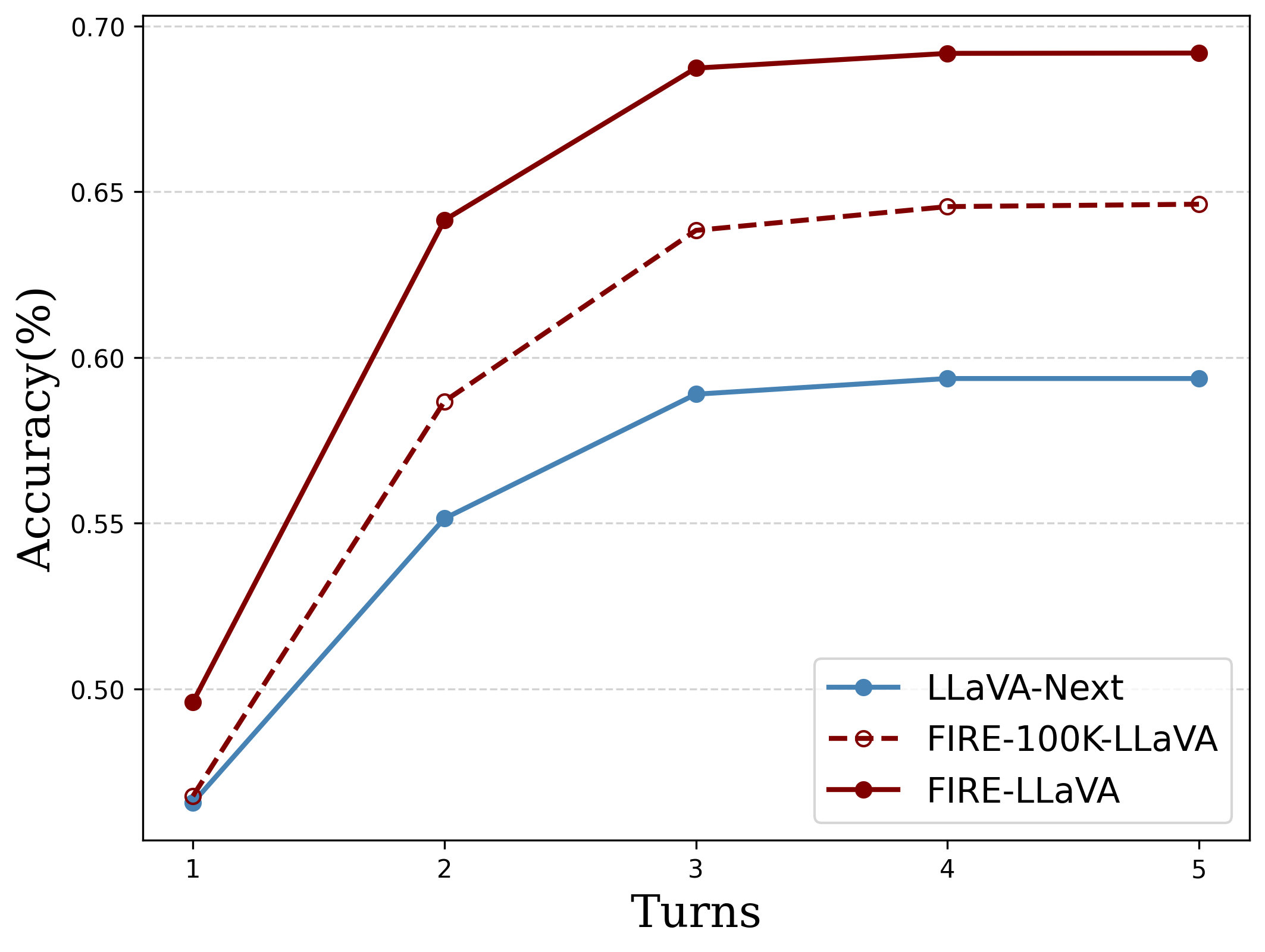}
    \caption{Accuracy improvement with more dialogue turns.}
    \label{fig:supp-acc-turn}
\end{figure}

In~\cref{fig:supp-acc-turn}, we present the performance curve in FIRE-Bench concerning the number of turns in dialogues, evaluating LLaVA-NeXT, \studentmodel, and \model. 
We report the percentage of correctly answered samples (those scores greater than 8) after each turn. 
As the number of turns increases, the percentage of correctly answered samples rises across all three models. $46.57\%$ and $46.77\%$ of the test data is correctly answered in the first turn, for the  LLaVA-NeXT model and \studentmodel respectively.
For \model, $49.60\%$ of the data is correctly answered in the first turn, and this increases to $69.19\%$ after five turns, with $19.59\%$ of the samples being corrected based on feedback.
Compared to the LLaVA-NeXT model, \model shows an additional $6.79\%$ improvement (from $49.60\% \to 69.19\%$ \emph{vs.} $46.57\% \to 59.37\%$), highlighting the effectiveness of \model when trained on FIRE.


\vskip -0.1in
\subsection{Visualization}
\begin{figure}[h]
    \vskip -0.1in
    \centering
    \includegraphics[width=1\linewidth]{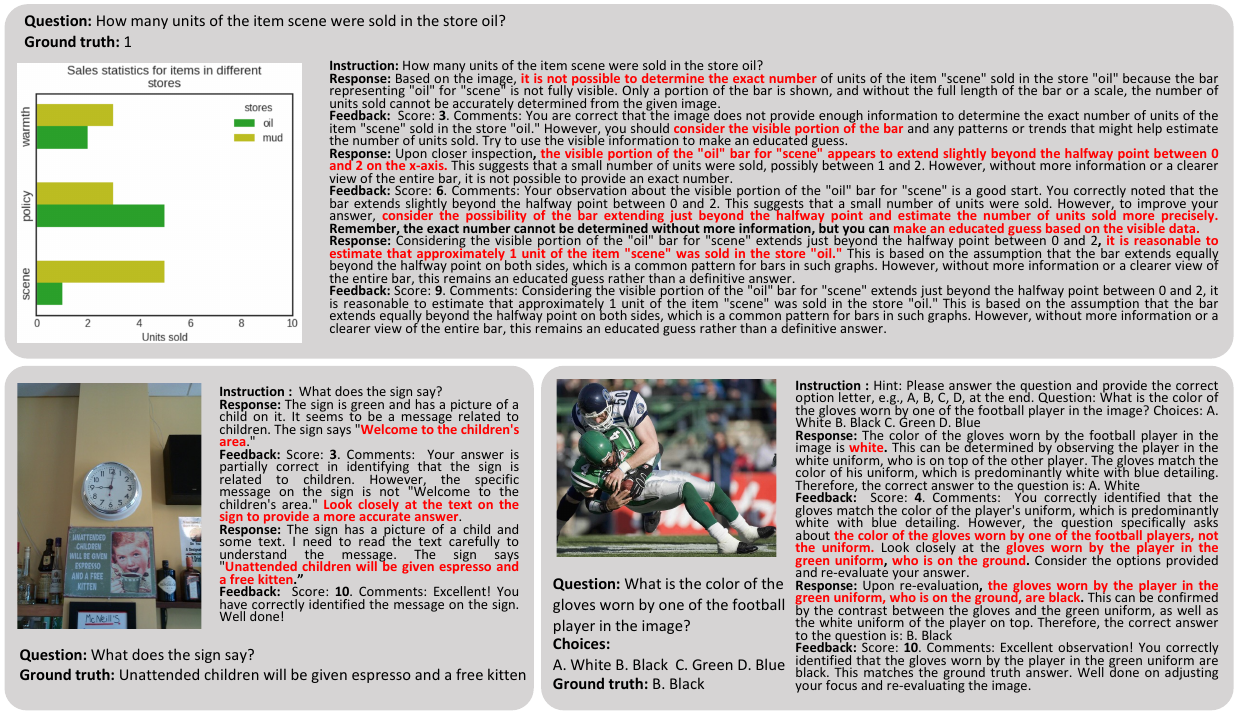}
    \vskip -0.1in
    \caption{Case study of the feedback-refining ability in our model.}
    \label{fig:visualization}
    \vskip -0.2in
\end{figure}

\vskip -0.1in
In~\cref{fig:visualization}, we visualize three cases in free dialogues using \model and \feedbackmodel. 
We observe that \model can understand the intent behind the feedback and refines its responses according to the feedback on chart understanding, visual concept perception, visual relationship reasoning, and OCR, making efficient user-agent interactions.

\vskip -0.2in
\section{Conclusion}
\vskip -0.15in
\label{sec:conclusion}
In this paper, we have presented FIRE, a feedback-refinement dataset with 1.1M multi-turn conversations, which empowers VLMs to refine their responses based on given feedback. 
Given proper prompts, GPT-4V can produce high-quantity conversations with feedback and responses.
Using the 100K GPT-4V generated data as seeds, a student model and a teacher model can freely expand the feedback-refinement data to 1.1M with a similar data quality to GPT-4V.
Experiments show that VLMs trained on FIRE have significant improvements in their feedback-refining ability.

\textbf{Limitation.}
In the current FIRE dataset, the feedback data is limited in the textual form. Practical feedback usually involves diverse multimodal information, such as pointing out image regions.
We will further expand FIRE with multimodal feedback data. In addition, although we use a filter process to remove low-quality data, we still cannot completely guarantee the quality of the data. In the future, we will combine human verification with machine verification to improve the quality.

\noindent \textbf{Acknowledgements.}
This work was partly supported by the National Science and Technology Major Project (2022ZD0114900). 
This work was partly supported by the Natural Science Foundation of China (NSFC) under Grants No. 62176021 and No. 62172041, the Natural Science Foundation of Shenzhen under Grant No. JCYJ20230807142703006, and the Key Research Platforms and Projects of the Guangdong Provincial Department of Education under Grant No.2023ZDZX1034. Mehrtash Harandi is supported by funding from the Australian Research Council Discovery Program DP230101176.

{
\small
\bibliographystyle{plain}
\bibliography{reference_header,references}
}

\clearpage
\appendix

\renewcommand\thefigure{A\arabic{figure}}
\setcounter{figure}{0}
\renewcommand\thetable{A\arabic{table}}
\setcounter{table}{0}
\renewcommand\theequation{A\arabic{equation}}
\setcounter{equation}{0}
\pagenumbering{arabic}
\renewcommand*{\thepage}{A\arabic{page}}
\setcounter{footnote}{0}

\section{Human Verification on FIRE}
\subsection{Human verification on data quality}
\begin{figure}[htbp]
    \centering
    \includegraphics[width=0.999\textwidth]{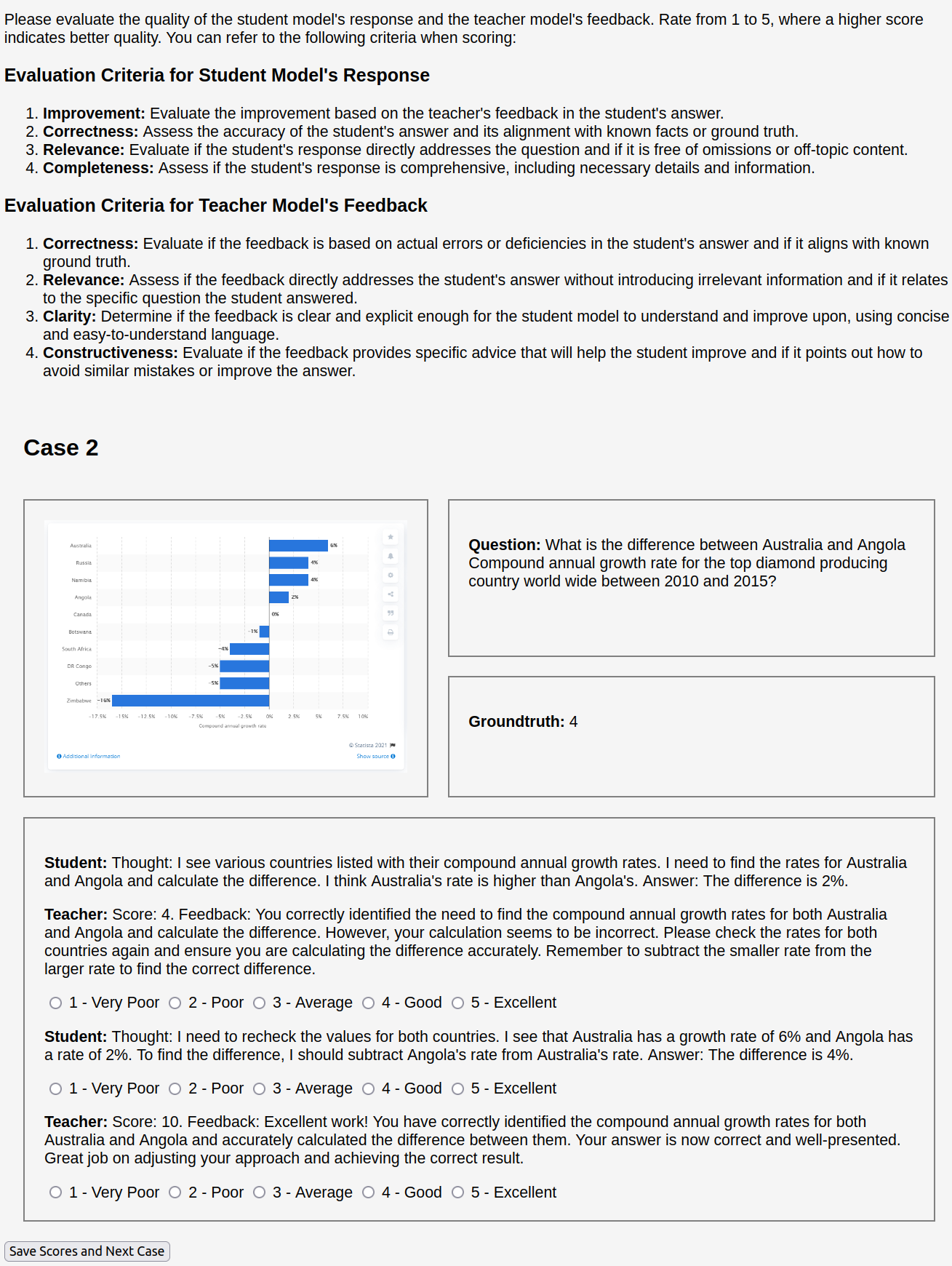}
    \caption{The screenshot of the interface for the human verification on data quality.}
    \label{fig:human-rating}
\end{figure}

To evaluate the data quality of generated data in FIRE-100K, FIRE-1M, and FIRE-Bench, we conduct a user study for the three splits of FIRE. Concretely, we randomly sample 100 conversations from each of the three splits, and ask 10 persons to provide scores (1-5) for feedback and refined responses in each turn of conversations.
For the feedback, we ask the person ``Please consider the quality of the refined feedback, based on its correctness, relevance, clarity, and constructiveness. Give a score (1-5). 1 means its quality is bad, and 5 means its quality is very good".
For the refined response, we ask the person ``Please consider the quality of the response, based on its improvement, correctness, and completeness. Given a score (1-5). 1 means its quality is bad, and 5 means its quality is very good".
The interface of the user study is shown in ~\cref{fig:human-rating}.
We report the average scores in~\cref{table:human_verification}.
We can find that, most users provide high scores for generated data in the three splits, showing that our dataset has high-quality data.

\begin{table}[htbp]
\vskip -0.15in
\caption{Average scores from humans on FIRE-100K, FIRE-1M, and FIRE-Bench, with 5 being the highest score.}
\vskip -0.05in
\label{table:human_verification}
\centering
\begin{tabular}{ c c | c c | c c}
    \bottomrule
    \multicolumn{2}{c|}{FIRE-100K} & \multicolumn{2}{c|}{FIRE-1M} & \multicolumn{2}{c}{FIRE-Bench} \\
    Feedback & Response & Feedback & Response & Feedback & Response \\ \hline
    4.87 & 4.66 & 4.84 & 4.73 & 4.88  & 4.74 \\
    \bottomrule
\end{tabular}
\vskip -0.05in
\end{table}

\subsection{Human verification on FIRE-LLaVA}

\begin{figure}
    \centering
    \fbox{\includegraphics[width=0.93\linewidth]{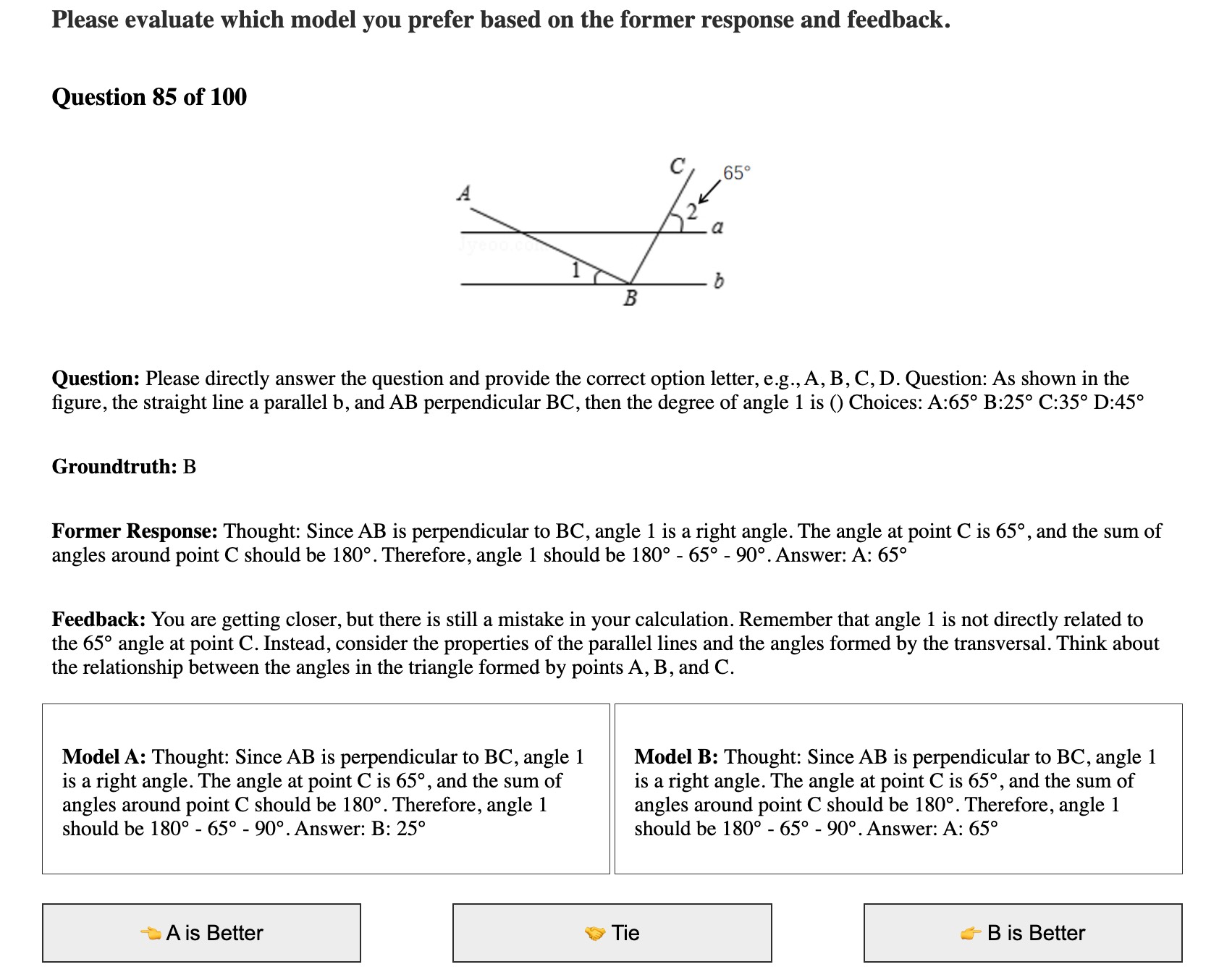}}
    \caption{The interface of human evaluation on FIRE LLaVA. We randomly sampled 100 instances, allowing both the baseline and FIRE-LLaVA models to generate inferences under identical former responses and feedback conditions. Three participants were invited to rate the responses. For each sample, the responses were shuffled to randomize the association between the responses and Model A/B. Participants selected their preference by choosing `A is better,’ `Tie,’ or `B is better’ based on their
judgment.}
    \label{fig:model_human_study}
\end{figure}

 To evaluate the models qualitatively, we conducted a human study comparing responses from LLaVA-Next-8B and FIRE-LLaVA. The interface is shown in \cref{fig:model_human_study}. We randomly sampled 100 instances and provided each model with identical initial responses and feedback, asking them to generate refined responses. Three independent human evaluators assessed these responses, without knowing which model generated which response (responses were shuffled to ensure blinding). The evaluation results, detailed in \cref{table:model_human_verification}, show that FIRE-LLaVA outperforms LLaVA-Next-8B with a significantly higher preference score (37.67 vs. 24.33), indicating that FIRE-LLaVA’s responses are more aligned with human preferences.

\begin{table}[htbp]
\caption{Human evaluation for LLaVA-Next-8B and FIRE-LLaVA.}
\label{table:model_human_verification}
\centering
\begin{tabular}{ l| c c c}
    \bottomrule
     & \textbf{FIRE-LLaVA is Better} & \textbf{Tie }&\textbf{LLaVA-Next-8B is Better} \\ \hline
    \textbf{Votes} & 37.67 & 38  & 24.33 \\
    \bottomrule
\end{tabular}
\end{table}

\section{Data source}
\label{appendix:data_source}
Our dataset uses images from 27 diverse sources to provide a robust training dataset for FIRE. All 27 datasets are public datasets, and all the images can be downloaded via links in \cref{tab:supp-datasource}. The comprehensive list of the source datasets and links to their metadata are detailed below:

\begin{table}[h]
\caption{Data utilized from 27 source datasets for training and test data in FIRE.}
\label{tab:supp-datasource}
\centering
\resizebox{1\columnwidth}{!}{
\begin{tabular}{llll  }
\toprule 
 \href{https://huggingface.co/datasets/liuhaotian/LLaVA-Instruct-150K}{LLaVA (train)}~\cite{llava} 
& \href{https://cocodataset.org/}{COCO (train)}~\cite{lin2014microsoft}
& \href{https://segment-anything.com/dataset/index.html}{SAM (train)}~\cite{kirillov2023segment} 
& \href{https://homes.cs.washington.edu/~ranjay/visualgenome/index.html}{VG (train)}~\cite{krishna2017visual} 
 \\ 

 \href{https://drive.google.com/drive/folders/1tCUQ-sq6vdshZVkF0ZeF3K4eztkXJgax}{Web-Landmark (train)}~\cite{chen2023sharegpt4v} 
& \href{https://drive.google.com/drive/folders/1tCUQ-sq6vdshZVkF0ZeF3K4eztkXJgax}{WikiArt (train)}~\cite{saleh2015large} 
& \href{https://ocr-vqa.github.io/}{OCRVQA (train)}~\cite{mishra2019ocr} 
& \href{https://allenai.org/data/diagrams}{AI2D (train)}~\cite{kembhavi2016diagram} 
  \\ 
 \href{https://github.com/FreedomIntelligence/ALLaVA}{ALLaVA-Vflan (train)}~\cite{chen2024allava} 
 & \href{https://drive.google.com/drive/folders/1tCUQ-sq6vdshZVkF0ZeF3K4eztkXJgax}{Web-Celebrity (train)}~\cite{chen2023sharegpt4v} 
 & \href{https://drive.google.com/drive/folders/1tCUQ-sq6vdshZVkF0ZeF3K4eztkXJgax}{Share-TextVQA (train)}~\cite{chen2023sharegpt4v}
 & \\ \hline

 \href{https://github.com/vis-nlp/ChartQA}{ChartQA (train\&test)}~\cite{masry2022chartqa} 
& \href{https://www.docvqa.org/}{DocVQA (train\&test)}~\cite{mathew2021docvqa} 
& \href{https://kushalkafle.com/projects/dvqa.html}{DVQA (train\&test)}~\cite{kafle2018dvqa} 
& \href{https://github.com/chen-judge/GeoQA}{GeoQA+ (train\&test)}\cite{chen2021geoqa} 
\\ 

 \href{https://visualqa.org/}{VQAV2 (train\&test)}~\cite{goyal2017making} 
& \href{https://cs.stanford.edu/people/dorarad/gqa/about.html}{GQA (train\&test)}~\cite{hudson2019gqa} & \href{https://textvqa.org/}{TextVQA (train\&test)}~\cite{singh2019towards} & \href{https://huggingface.co/datasets/naver-clova-ix/synthdog-en}{Synthdog-EN (train\&test)}\cite{kim2022ocr}   \\ \hline

\href{https://huggingface.co/datasets/liuhaotian/llava-bench-in-the-wild}{LLaVA-in-the-Wild (test)}~\cite{llava}
& \href{https://mmmu-benchmark.github.io/}{MMMU (test)}~\cite{yue2023mmmu} 
& \href{https://github.com/QwenLM/Qwen-VL/blob/master/eval_mm/mme/EVAL_MME.md}{MME (test)}~\cite{fu2024mme} 
& \href{https://github.com/yuweihao/MM-Vet}{MM-Vet (test)}~\cite{yu2023mm} 
 \\ 

\href{https://mathverse-cuhk.github.io/}{MathVerse (test)}~\cite{zhang2024mathverse} 
& \href{https://mathvista.github.io/}{MathVista (test)}~\cite{lu2023mathvista} 
&\href{https://scienceqa.github.io/}{ScienceQA (test)}~\cite{lu2022scienceqa}
& \href{https://github.com/AILab-CVC/SEED-Bench}{SEED-bench (test)}~\cite{li2023seed} \\

\bottomrule
\end{tabular}
}
\end{table}

\section{Additional Experimental Results}

\subsection{Error bar}
We report the error bar of average turn (AT), average dialogue refinement (ADR), average turn refinement (ATR), and refinement ratio (RR) in fixed dialogues. We run the model three times and compute the standard deviation, as shown in~\cref{table:error_bar}. 
Comparisons among the four metrics, the standard deviation is relatively small, less than $8\%$ of the average results, showing that our method can achieve stable feedback-refining ability. 
\begin{table}[htbp]
\caption{Results in free dialogue over all test data in FIRE.}
\vskip -0.1in
\label{table:error_bar}
\centering
\footnotesize
\begin{tabular}{ c |  c c c c }
    \toprule
   Model & AT ($\downarrow$) & ADR ($\uparrow$) & ATR ($\uparrow$) & RR ($\uparrow$) \\
   \hline
  LLaVA-Next-8B  & 1 &	0.97&	0.41&	0.25 \\
   \studentmodel-8B  &  0.92 $\pm$ 0.026 &	1.27 $\pm$ 0.013&	0.55 $\pm$ 0.042 &	0.34 $\pm$ 0.022 \\   
   \model-8B  &\textbf{0.84 $\pm$ 0.015} &	\textbf{1.56 $\pm$ 0.012}&	\textbf{0.66 $\pm$ 0.053}&	\textbf{0.39 $\pm$ 0.028} \\
    \bottomrule
\end{tabular}
  \vskip -0.2in
\end{table}

\subsection{More VLMs}

\subsubsection{LLaVA-Next-Vicuna-7B}
We further train a \vicunamodel model that replaces LLaMA3-8B in \model with Vicuna1.5-7B. Results are shown in~\cref{table:morevlms}. Results of using Vicuna1.5-7B demonstrate the effectiveness of FIRE again, where \vicunamodel has better feedback-refining ability than the original LLaVA-Next-Vicuna model on AT, ADR, ATR, and RR, showing the helpfulness for the feedback-refining ability.

\begin{table}
\caption{Results of \vicunamodel in free dialogue over all test data in FIRE.}
\vskip -0.1in
\label{table:morevlms}
\centering
\footnotesize
\begin{tabular}{ c |  c c c c }
    \toprule
   Model & AT ($\downarrow$) & ADR ($\uparrow$) & ATR ($\uparrow$) & RR ($\uparrow$) \\
   \hline
  LLaVA-Next-Vicuna  & 1.00 &	0.98&	0.49&	0.24 \\   
   \vicunamodel  & \textbf{0.94} &	\textbf{1.11} &	\textbf{0.57} &	 \textbf{0.27} \\
    \bottomrule
\end{tabular}
  \vskip -0.2in
\end{table}

\subsubsection{LLaVA-1.5-Vicuna-7B}
We have also performed experiments using LLaVA-1.5-Vicuna-7B as another baseline. The results are presented in \cref{table:morevlm-1.5}. The findings demonstrate that the LLaVA-1.5-Vicuna-7B model fine-tuned on FIRE100K outperforms the original LLaVA-1.5-Vicuna-7B model across all four metrics—AT, ADR, ATR, and RR—highlighting the effectiveness of our FIRE dataset.

\begin{table}
\caption{Results of LLaVA-1.5-Vicuna-7B and LLaVA-1.5-Vicuna-7B-FIRE on FIRE-Bench.}
\vskip -0.1in
\label{table:morevlm-1.5}
\centering
\footnotesize
\begin{tabular}{ c |  c c c c }
    \toprule
   Model & AT ($\downarrow$) & ADR ($\uparrow$) & ATR ($\uparrow$) & RR ($\uparrow$) \\
   \hline
  LLaVA-1.5-Vicuna  & 1.00 &	0.62&	0.46&	0.12 \\   
  FIRE-LLaVA-1.5  & \textbf{0.94} &	\textbf{0.80} &	\textbf{0.61} &	 \textbf{0.20} \\
    \bottomrule
\end{tabular}
\end{table}

\subsection{More Visualization}

\begin{figure}
    \centering
    \includegraphics[width=0.65\textwidth]{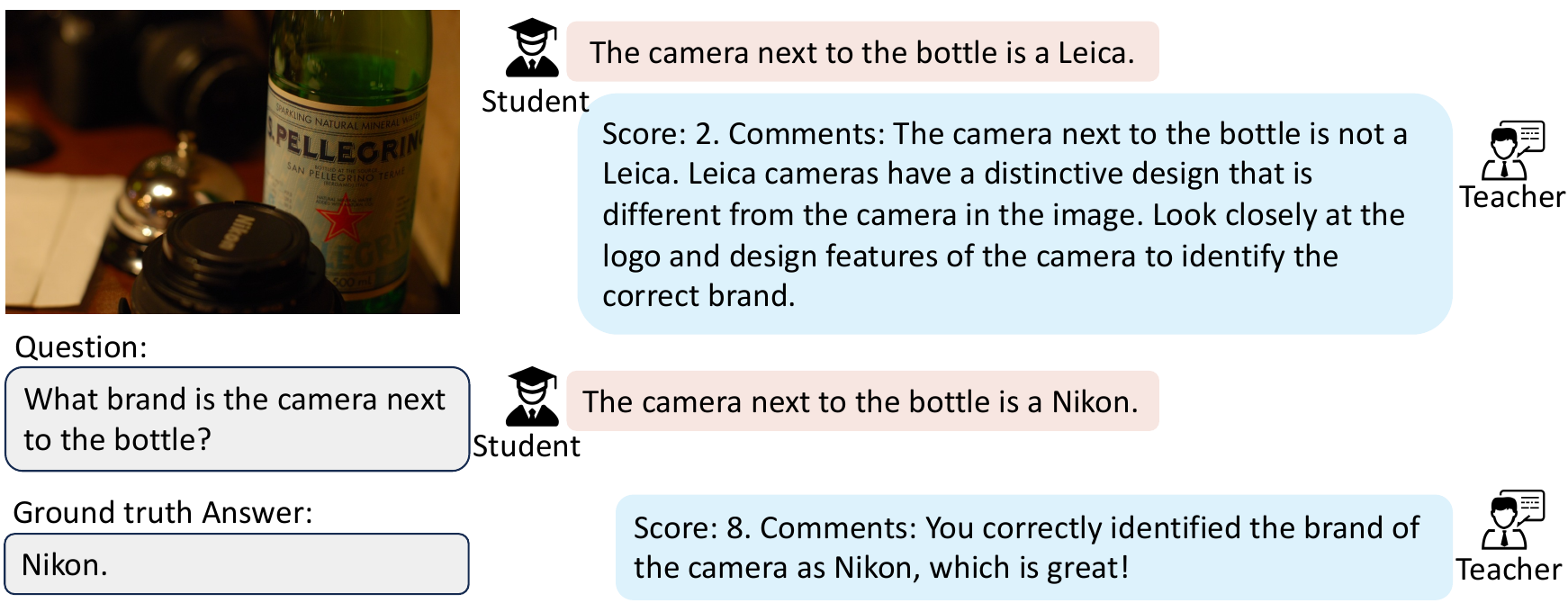}
    \caption{One case for feedback-refining ability in OCR.}
    \label{fig:result1}
\end{figure}

\begin{figure}
    \centering
    \includegraphics[width=0.999\textwidth]{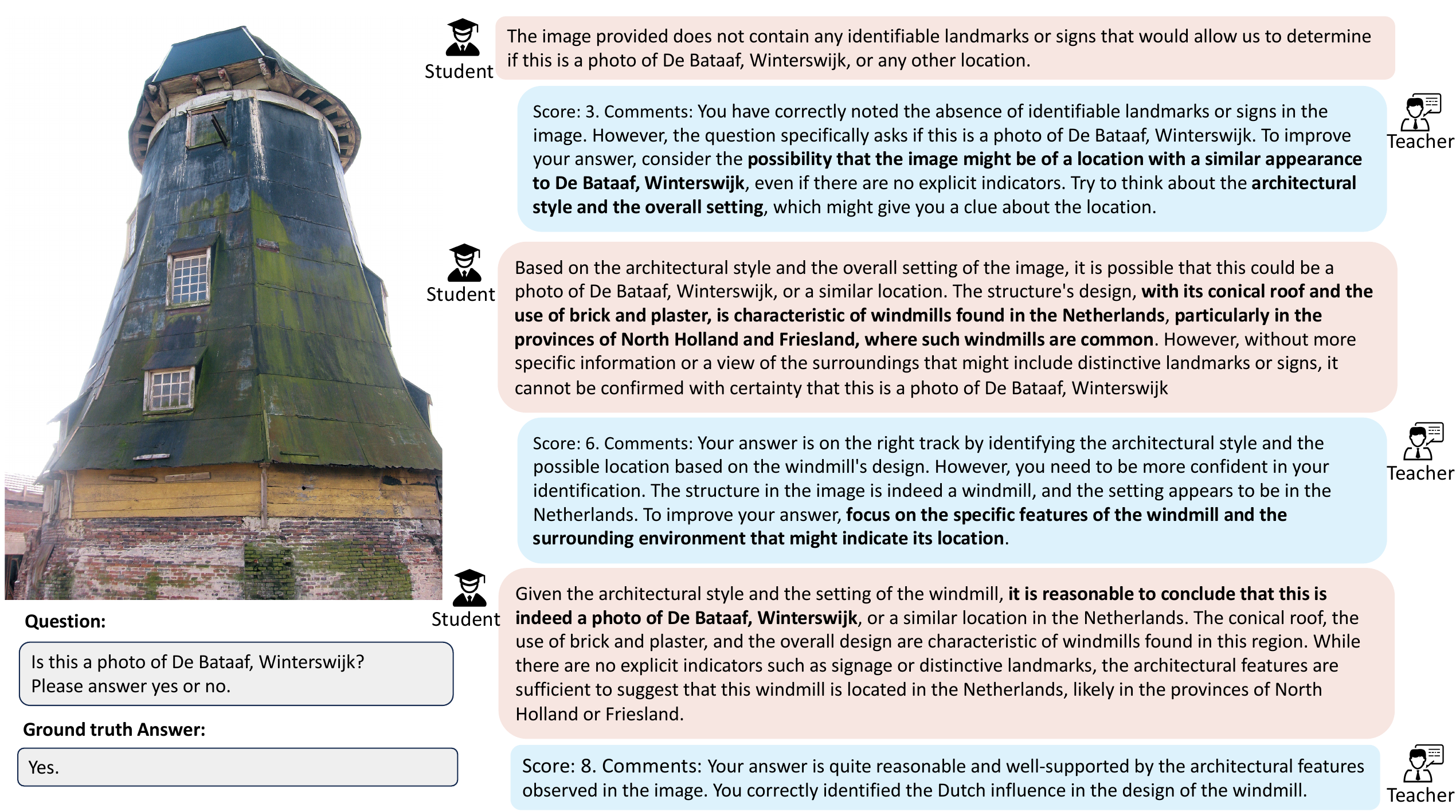}
    \caption{One case for feedback-refining ability in visual perception.}
    \label{fig:result2}
\end{figure}

\begin{figure}
    \centering
    \includegraphics[width=0.999\textwidth]{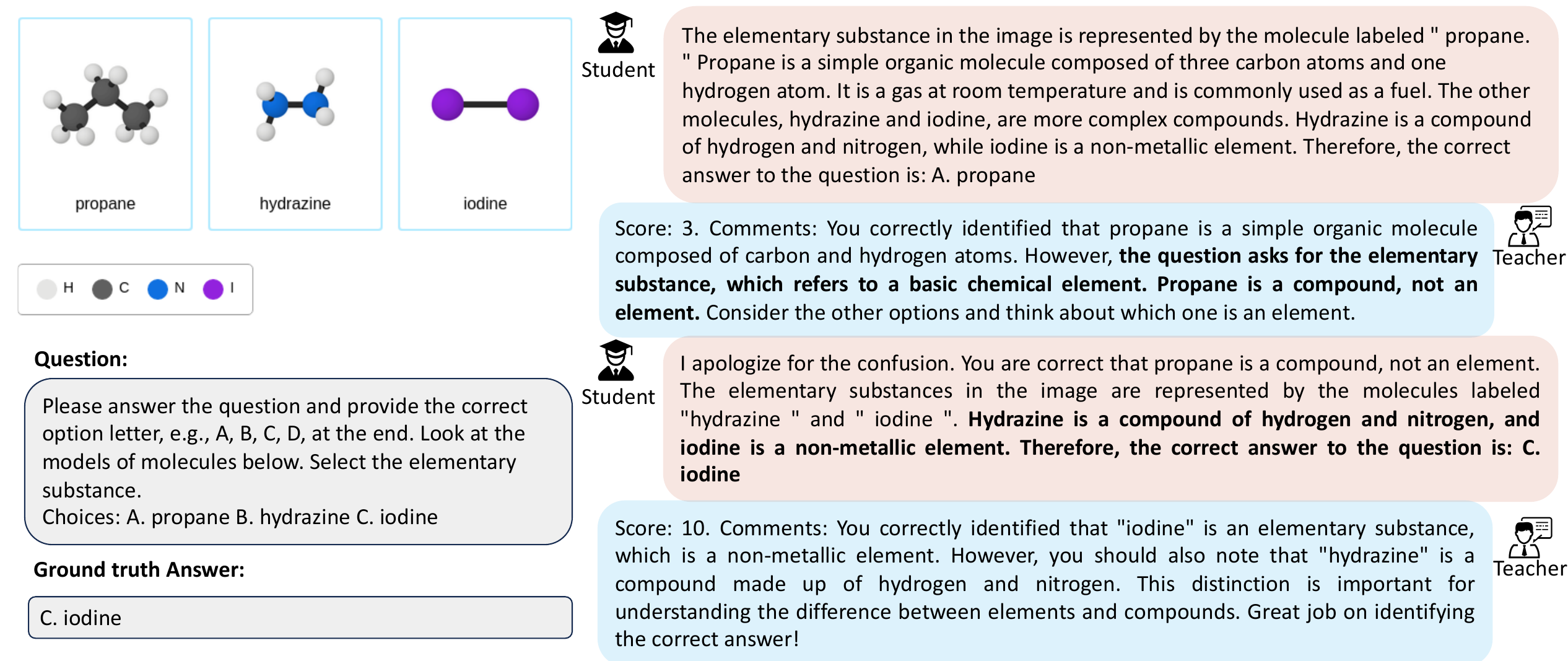}
    \caption{One case for feedback-refining ability in science question answering.}
    \label{fig:result3}
\end{figure}

\begin{figure}
    \centering
    \includegraphics[width=0.999\textwidth]{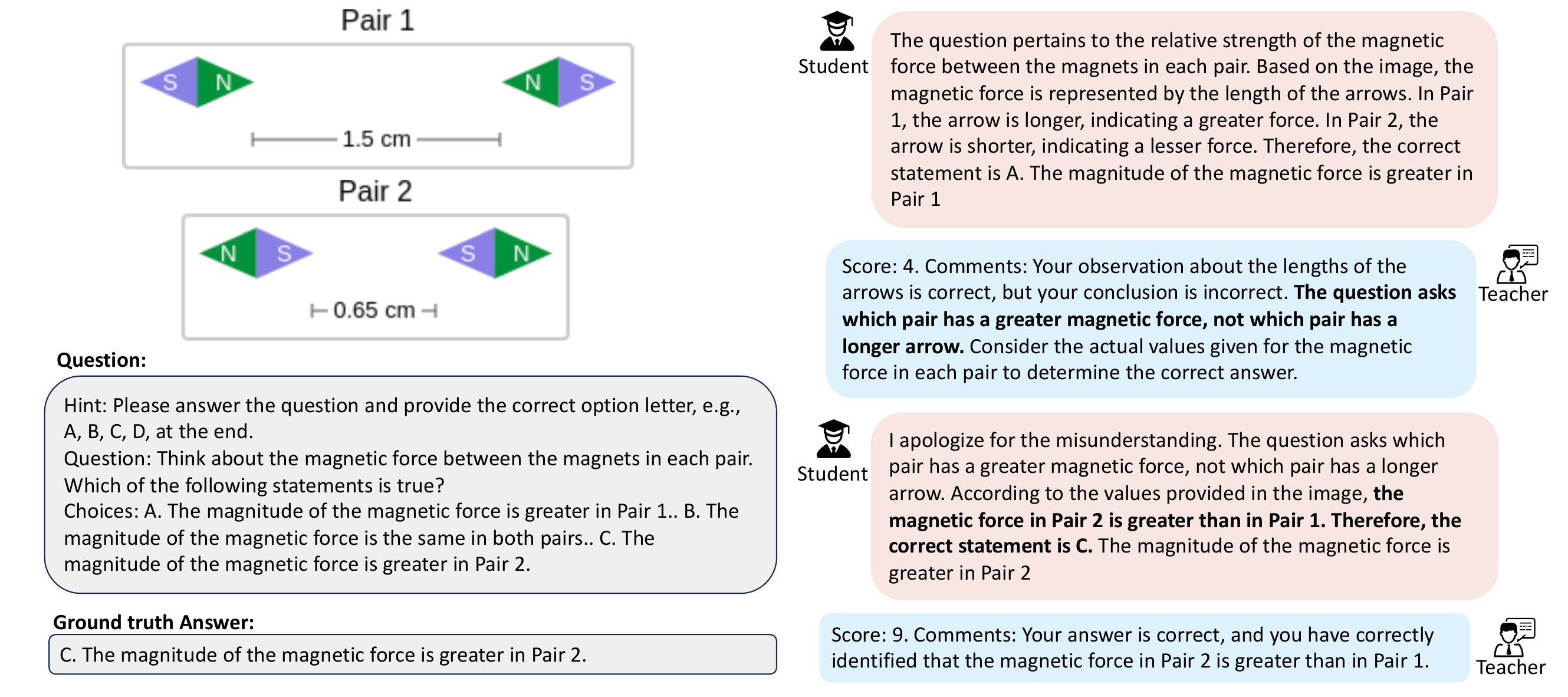}
    \caption{One case for feedback-refining ability in science question answering.}
    \label{fig:result4}
\end{figure}

\begin{figure}
    \centering
    \includegraphics[width=0.999\textwidth]{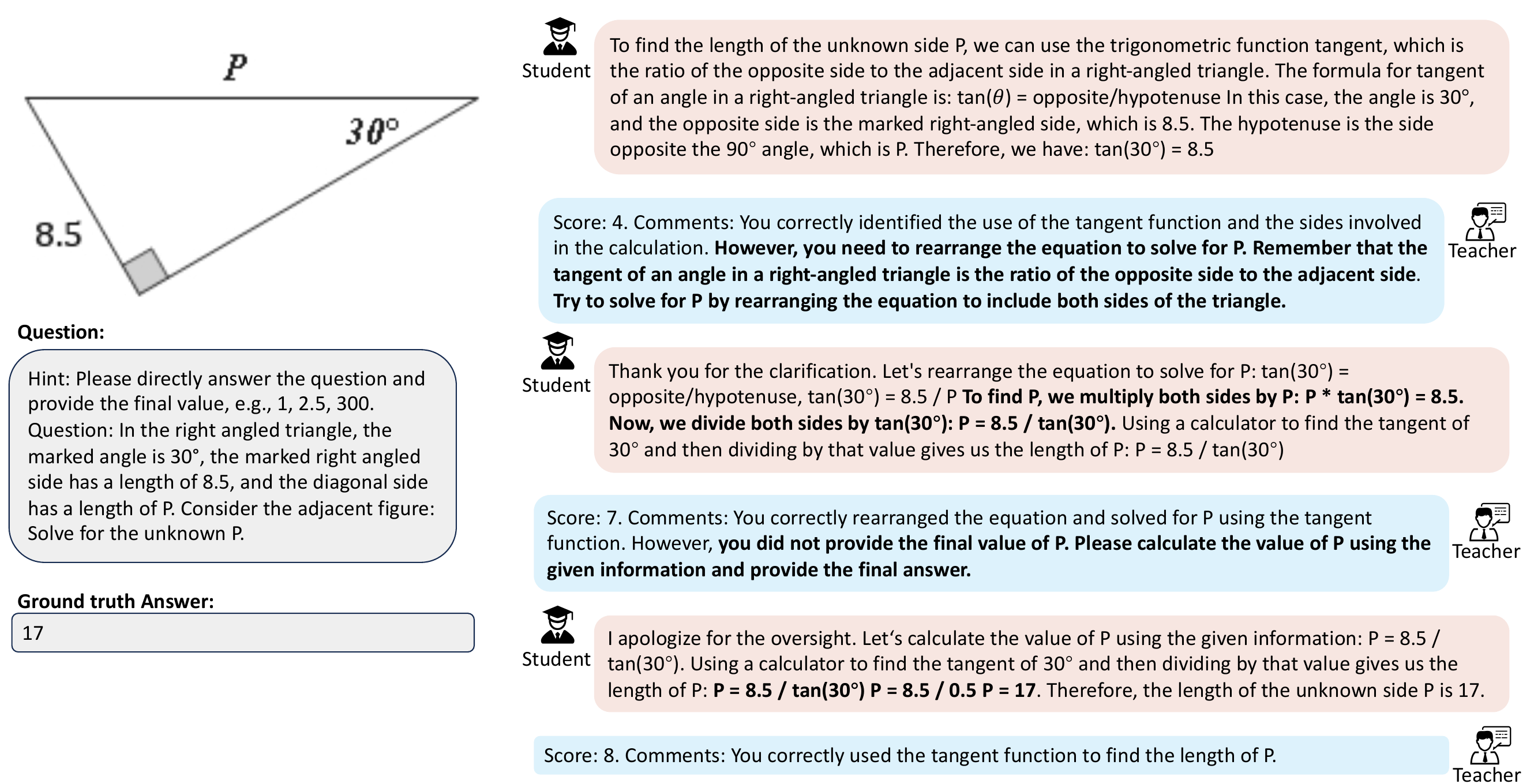}
    \caption{One case for feedback-refining ability in math reasoning.}
    \label{fig:result5}
\end{figure}

\begin{figure}
    \centering
    \includegraphics[width=0.999\textwidth]{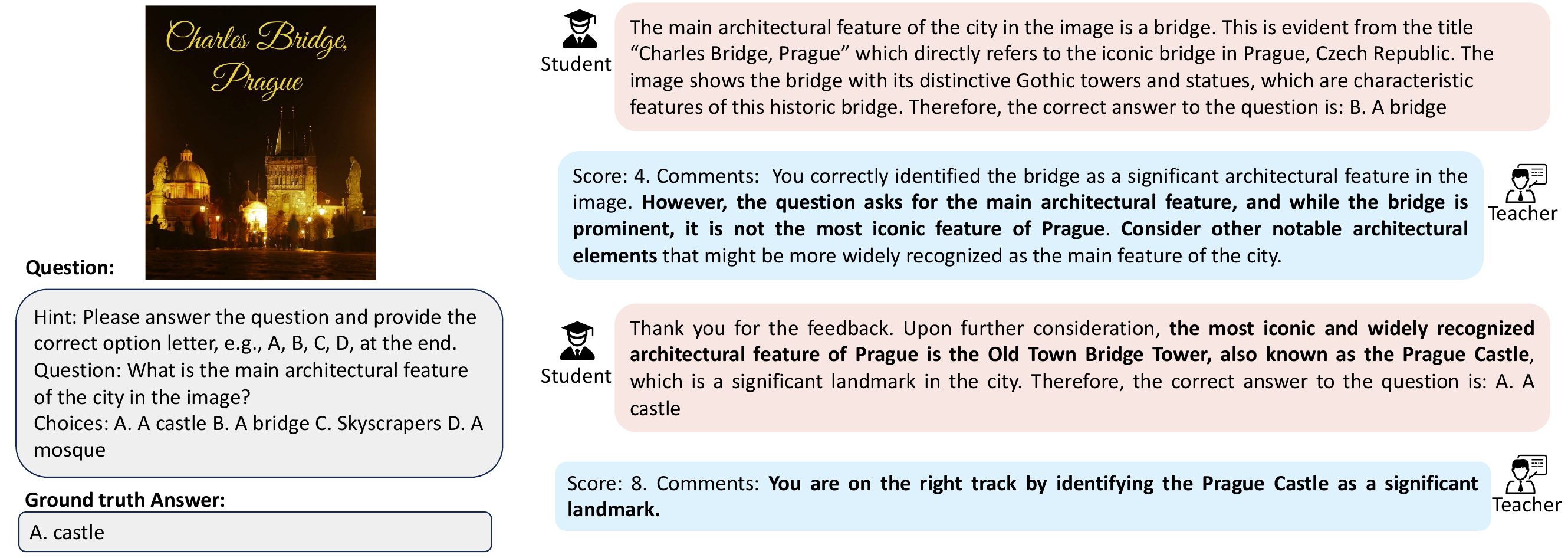}
    \caption{One case for feedback-refining ability in visual question answering.}
    \label{fig:result6}
\end{figure}

\begin{figure}
    \centering
    \includegraphics[width=0.999\textwidth]{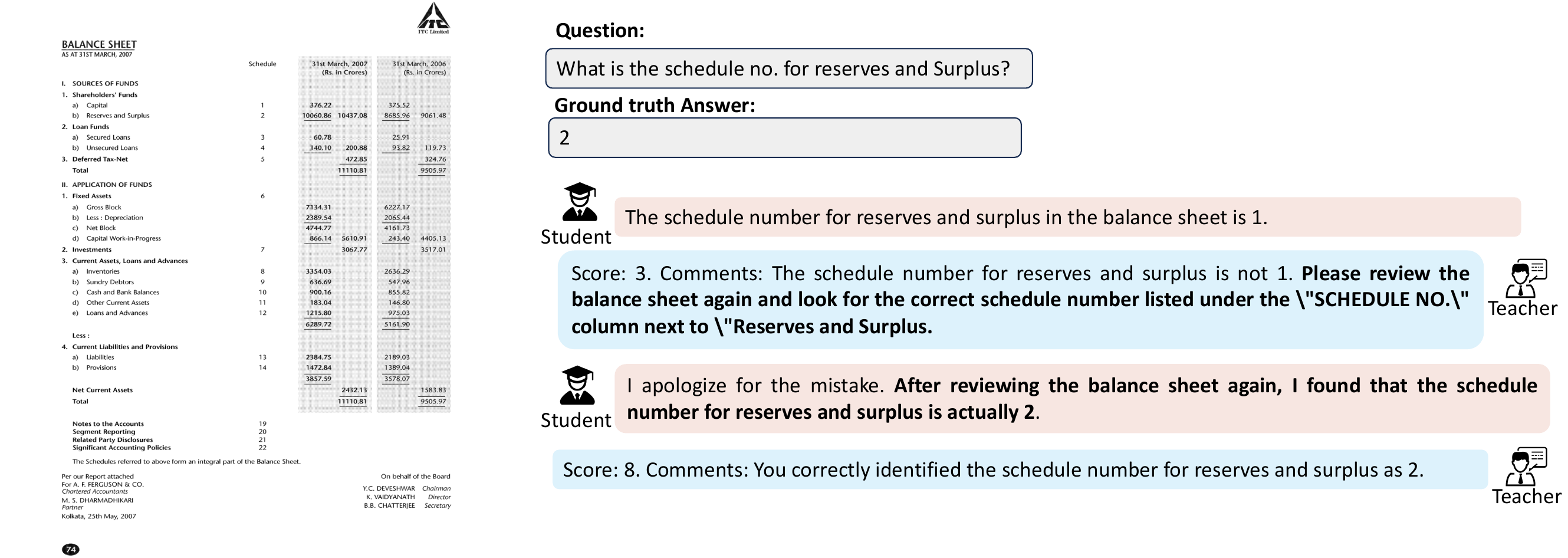}
    \caption{One case for feedback-refining ability in document understanding.}
    \label{fig:result7}
\end{figure}

\begin{figure}
    \centering
    \includegraphics[width=0.999\textwidth]{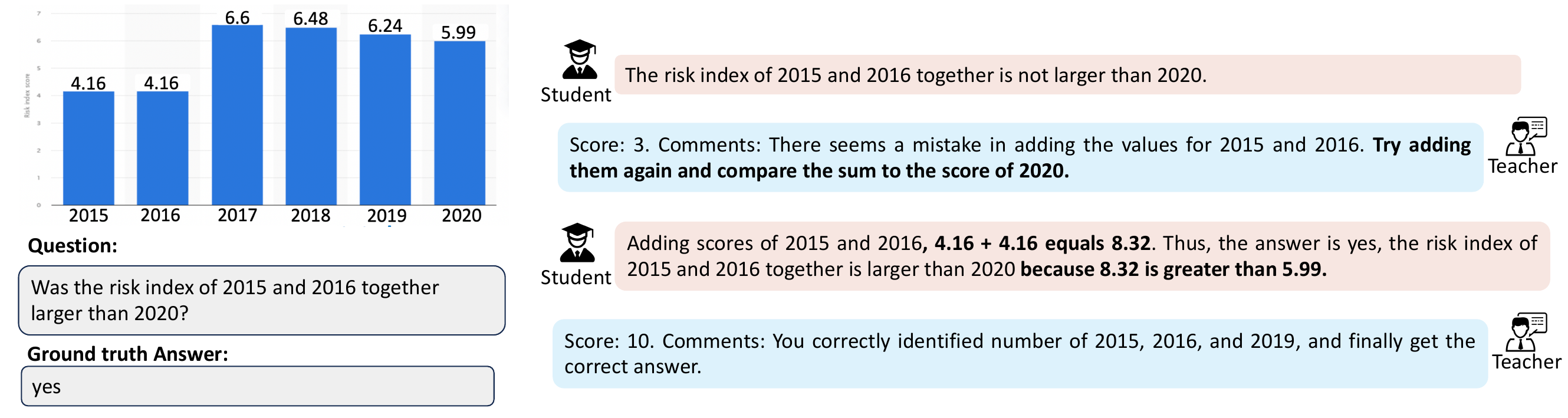}
    \caption{One case for feedback-refining ability in chart analysis.}
    \label{fig:result8}
\end{figure}

We provide more cases tested on the FIRE-Bench to show the feedback-refining ability, as shown in~\cref{fig:result1,fig:result2,fig:result3,fig:result4,fig:result5,fig:result6,fig:result7,fig:result8}.
We observe that \model trained on FIRE has good feedback-refining ability on different tasks, such as visual perception, visual question answering, image captioning, OCR reasoning, document understanding, math reasoning, science question answering, and chart analysis.

\section{Prompt}

\subsection{Prompt for GPT-4V}
We show examples of prompts for GPT-4V to generate feedback-refinement conversations, as shown in ~\cref{fig:gpt_prompt_for_system} and ~\cref{fig:gpt_prompt_for_user}.

\begin{figure}[htbp]

\begin{minipage}{0.99\columnwidth}\vspace{0mm}    \centering
\begin{tcolorbox}
\fontsize{9.0pt}{\baselineskip}\selectfont
You are a helpful assistant that can generate a dialogue between a teacher and a student. The student is trying to answer a question about an image. The student first gives a wrong answer. Based on the given groundtruth answer, the teacher provides feedback to help the student gradually improve its answer.
\\
Use the following template to generate the dialogue:\\ \\
"""\\
\# \textbf{Round 1}\\ \\
\#\# \textbf{Student's response}\\
{Thought:} <here is the student's thought process about the question. Do NOT use the words "teacher" or "student". >\\
{Answer:} <here is the student's answer to the question.>\\

\#\# \textbf{Teacher's feedback}\\
{Score:} <compare the student's answer with the groundtruth answer in terms of accuracy, relevance, helpfulness, and level of detail. Provide an overall score on a scale of 1 to 10, where a higher score indicates better overall performance.>\\
{Feedback:} <provide feedback on the student's answer. Do not directly tell the groundtruth answer. The feedback should identify which parts of the student's answer are incorrect, what is missing in the student's answer, and how to improve the student's answer.>\\

\# \textbf{Round 2}\\
    \indent ...\\

\# \textbf{Round n}\\
    \indent ...\\

"""\\
The number of rounds should depend on the difficulty of the question. More rounds should be used for difficult questions, while fewer rounds should be used for easy questions.

\end{tcolorbox}
\caption{System prompt for GPT-4V for Student-Teacher conversation generation.}
\label{fig:gpt_prompt_for_system}
\end{minipage}
\end{figure}

\begin{figure*}[htbp]
\begin{minipage}{0.99\columnwidth}\vspace{0mm}    \centering
\begin{tcolorbox}
\fontsize{9.0pt}{\baselineskip}\selectfont
Here are the given image, question: \blueprompt{<question>} and groundtruth answer: \blueprompt{<groundtruth>}, now generate a dialogue:
\end{tcolorbox}
\caption{User prompt for GPT-4V for Student-Teacher conversation generation.}
\label{fig:gpt_prompt_for_user}
\end{minipage}
\end{figure*}

\subsection{Prompt for Student and Teacher models}
We show examples of prompts for student and teacher models to simulate feedback-refinement conversations, as shown in \cref{fig:prompt_for_student_model} and \cref{fig:prompt_for_teacher_model}, respectively. In \cref{fig:prompt_for_student_model}, the prompt for the student model to generate $n$-th response is shown. The prompt contains the last $n-1$ rounds' student responses and the textual comments from the teacher model. The prompt for the teacher model is shown in \cref{fig:prompt_for_teacher_model}.  Firstly, the prompt provides user instruction and ground truth. Secondly, the prompt contains instructions that format the teacher model's feedback as textual comments and numeric scores. Finally, the prompt incorporates the only latest student response into its context.
\begin{figure*}[htbp]
\begin{minipage}{0.99\columnwidth}\vspace{0mm}    \centering
\begin{tcolorbox}
\fontsize{9.0pt}{\baselineskip}\selectfont
You are a helpful language and vision assistant. You are able to understand the visual content that the user provides, and assist the user with a variety of tasks using natural language
\\ 
\blueprompt{<user\_instruction>}
\\  \\
\# \textbf{Round 1} \\
\blueprompt{<student\_response\_round\_1>}\\
\blueprompt{<feedback\_round\_1>}\\ \\
...\\
\# \textbf{Round n-1} \\
\blueprompt{<student\_response\_round\_n-1>}\\
\blueprompt{<feedback\_round\_n-1>}\\ 

Based on the feedback, answer the question again:
\end{tcolorbox}
\caption{Prompt for student model to simulate feedback-refinement conversations.}
\label{fig:prompt_for_student_model}
\end{minipage}
\end{figure*}

\begin{figure*}[htbp]
\begin{minipage}{0.99\columnwidth}\vspace{0mm}    \centering
\begin{tcolorbox}
\fontsize{9.0pt}{\baselineskip}\selectfont

You are a helpful language and vision assistant. You are able to understand the visual content that the user provides, and assist the user with a variety of tasks using natural language\\

Question: \blueprompt{<question>} \\
Groundtruth: \blueprompt{<groundtruth>} \\ \\
Please compare my answer with the groundtruth answer and provide helpful, detailed, and polite feedback to help me improve my answer.
Formulate the feedback as: \\
"""\\
Score: <compare the provided response with the groundtruth answer in terms of accuracy, relevance, helpfulness, and level of detail, and provide an overall score on a scale of 1 to 10, where a higher score indicates better overall performance.> \\ \\
Feedback: <provide feedback on the response. Do NOT directly tell the groundtruth answer. The feedback should identify which parts of my answer are incorrect, what is missing in the response, and how to improve the response.>
\\"""\\
Here is the student response: \blueprompt{<student\_response>}, now please provide the feedback:
\end{tcolorbox}
\caption{Prompt for teacher model to simulate feedback-refinement conversations.}
\label{fig:prompt_for_teacher_model}
\end{minipage}
\end{figure*}

\section {Potential Negative Societal Impacts}

The VLMs may generate harmful outputs based on human induction feedback, resulting in risks, such as false information, discrimination, violent and pornographic content, and privacy leaks \emph{etc}.
To mitigate the risks of these harmful outputs, we will strictly filter and review the model outputs based on feedback in the future. 
In addition, users may become overly dependent on the model's outputs given feedback, neglecting the need for independent thinking and verification of information.



\end{document}